\documentclass[journal]{IEEEtran}
\usepackage{amsmath,amsfonts}
\usepackage{algorithmic}
\usepackage{algorithm}
\usepackage{array}
\usepackage[caption=false,font=scriptsize,labelfont=rm,textfont=sf]{subfig}
\usepackage{textcomp}
\usepackage{stfloats}
\usepackage{url}
\usepackage{verbatim}
\usepackage{graphicx}
\usepackage{multirow} 
\usepackage{color}
\usepackage{bbding} 
\usepackage{pifont}
\usepackage{bm}
\begin{document}
\title{Learning to Learn Transferable Generative Attack for Person Re-Identification}

\author{Yuan Bian, Min Liu, Xueping Wang, Yunfeng Ma, and Yaonan Wang
  % <-this % stops a space 
  \thanks{This work was supported in part by the National Natural Science Foundation of China under Grant 62221002, 62425305 and U22B2050, in part by the Science and Technology Innovation Program of Hunan Province under Grant 2023RC1048, in part by the Hunan Provincial Natural Science Foundation of China under Grant 2024JJ3013, in part by the Hunan Provincial Innovation Foundation for Postgraduate under Grant QL20230098. (\textit{Corresponding author: Min Liu}) }% <-this % stops a space   
  \thanks{
    Yuan Bian, Min Liu, Yunfeng Ma, and Yaonan Wang are with the College of Electrical and Information Engineering at Hunan University and National Engineering Research Center of Robot Visual Perception and Control Technology, Changsha, Hunan, China. E-mail: yuanbian$@$hnu.edu.cn; liu\_min$@$hnu.edu.cn; ismyf$@$hnu.edu.cn; yaonan$@$hnu.edu.cn.

    Xueping Wang is with the College of Information Science and Engineering at Hunan Normal University, and Hunan Provincial Key Laboratory of Intelligent Computing and Language Information Processing, Changsha, Hunan, China. E-mail: wang\_xueping$@$hnu.edu.cn.
  }
}


% \textit{i.e.}EEpubid{0000--0000/00\$00.00~\copyright~2021 IEEE}
% Remember, if you use this you must call \textit{i.e.}EEpubidadjcol in the second

\maketitle

\begin{abstract}
  Deep learning-based person re-identification (re-id) models are widely employed in surveillance systems and inevitably inherit the vulnerability of deep networks to adversarial attacks. Existing attacks merely consider cross-dataset and cross-model transferability, ignoring the cross-test capability to perturb models trained in different domains. To powerfully examine the robustness of real-world re-id models, the Meta Transferable Generative Attack (MTGA) method is proposed, which adopts meta-learning optimization to promote the generative attacker producing highly transferable adversarial examples by learning comprehensively simulated transfer-based cross-model\&dataset\&test black-box meta attack tasks. Specifically, cross-model\&dataset black-box attack tasks are first mimicked by selecting different re-id models and datasets for meta-train and meta-test attack processes. As different models may focus on different feature regions, the Perturbation Random Erasing module is further devised to prevent the attacker from learning to only corrupt model-specific features. To boost the attacker learning to possess cross-test transferability, the Normalization Mix strategy is introduced to imitate diverse feature embedding spaces by mixing multi-domain statistics of target models.  
  Extensive experiments show the superiority of MTGA, especially in cross-model\&dataset and cross-model\&dataset\&test attacks, our MTGA outperforms the SOTA methods by 20.0\% and 11.3\% on mean mAP drop rate, respectively. The source codes are available at \url{https://github.com/yuanbianGit/MTGA}.
\end{abstract}


\section{Introduction}
\IEEEPARstart{P}{erson} re-identification aims at retrieving specific persons from security surveillance systems \cite{ye2021deep,zheng2016person}. \textcolor{black}{Along with the advancement of deep learning, it has made remarkable progresses and been widely applied to intelligent surveillance systems \cite{huang2024meta,yang2023win,zhang2023multi,yan2023clip, wang2021learning,Li2020Multi,liu2023weakly,ahmed2015improved,wang2018cascaded,he2021transreid,bian2023occlusion}. However, it has been found that deep neural networks are vulnerable to adversarial attacks \cite{szegedy2014intriguing,goodfellow2014explaining,moosavi2017universal, Zhu2022Toward,Wang2021Universal,singla2020second,kumar2023impact,chen2024content,chen2024diffusion,li2023towards}, which can mislead deep neural network models by adding imperceptible perturbations to benign images. Deep learning-based re-id models inevitably inherit the vulnerability of deep networks to adversarial samples \cite{bai2020adversarial,wang2019advpattern}, which makes public safety under great threat.}
To study the security of surveillance systems, it is important to explore the vulnerability of the deep learning-based re-id models to adversarial samples.

\begin{figure}[t]
  \centering
  \subfloat[Black-box cross-model attack on classification tasks.]{
    \includegraphics[width=0.95\linewidth]{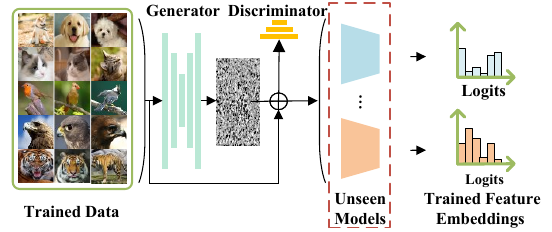}
    \label{fig1-a}
  }
  \quad
  \vspace{0.1cm}
  \centering
  \subfloat[Black-box cross-model, cross-dataset and cross-test attack on re-id tasks.]{
    \includegraphics[width=0.95\linewidth]{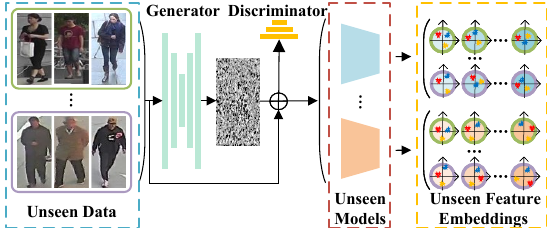}
    \label{fig1-b}
  }
  \quad
  \centering
  \subfloat{
    \includegraphics[width=0.95\linewidth]{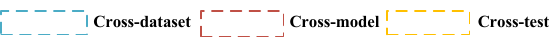}
    \label{fig1-c}
  }
  \vspace{-0.1cm}
  \textcolor{black}{\caption{Comparison of transfer-based black-box generative attacks between classification and re-id tasks. In black-box attack on classification tasks, the target models share the same feature embedding space and the training data of these models are aimed to be attacked. In black-box attack on re-id tasks, the target models may have diverse feature embedding spaces and unseen domain queries need to be attacked. Therefore, the re-id task attack has additional cross-dataset and cross-test transferability demands compared to the cross-model demand with the classification task attack.}}
  \label{fig:fig-1}
  % \vspace{-0.3cm}
\end{figure}

Recently, some works \cite{bai2020adversarial,wang2019advpattern,zheng2023u,bouniot2020vulnerability} have demonstrated that re-id models are susceptible to adversarial examples and introduced white-box adversarial metric attack methods to attack re-id models. These methods are not suitable in realistic scenarios, where parameters of target re-id models are not accessible. Transferable adversarial examples against black-box re-id models are then studied \cite{yang2022towards,yang2021learning,ding2021beyond,wang2020transferable,subramanyam2023meta}.
Different from transfer-based black-box attacks for classification tasks, which assume attackers have access to the training data of target model and generally only consider cross-model transferability among models trained in the same data distribution \cite{li2023cdta,zhang2021beyond}, attacks on black-box re-id models are more challenging due to the cross-model (architecture discrepancy between surrogate model and target model), cross-dataset (domain discrepancy between training image and target image) and cross-test
(domain discrepancy between target image and target model) transfer capabilities  are supposed, like Fig. \ref{fig:fig-1} shows. 
Specifically, re-id is an open-set task \cite{panareda2017open,gong2020adversarial}, where identities in the training and testing sets are non-overlapped and unseen query images often encounter a large domain shift \cite{zhong2018camera}, thus cross-dataset transferability is necessary for black-box adversarial attacks against re-id models. 
Except for cross-model transferability to attack models with different architectures, cross-test capability should take into account to attack models with different feature embedding spaces, since target re-id models could be trained with arbitrary domain datasets.
However, existing transfer-based re-id attacks do not fully consider these aspects, either ignoring cross-dataset capabilities \cite{ding2021beyond,wang2020transferable} or merely focusing on cross-model transferability and neglecting the cross-test capabilities \cite{yang2022towards,yang2021learning,subramanyam2023meta}, which leads to insufficient transferability of generated adversarial samples to effectively test the robustness of real-world re-id models.

In order to generate highly transferable adversarial examples against person re-id models, we propose the Meta Transferable Generative Attack (MTGA) approach, which utilizes meta-learning optimization to guide the generative attacker possessing the generic transferability by learning multiple simulated cross-model\&dataset\&test black-box meta attack tasks. Various train-test processes of cross-model\&dataset transfer-based black-box attacks are first generated as meta-learning tasks by Cross-model\&dataset Attack Simulation (CAS) method.
In terms of cross-dataset mimicking, multi-source datasets in the data zoo are utilized to randomly represent the adversarial attack training data and unseen domain testing data. For cross-model imitation, the agent model and the target model are picked differently in model zoo, which consists of three classical re-id models that can well represent global-based, part-based and attention-based approaches, considering these three types of re-id methods are most widely applied. 
Besides, considering limited surrogate model resources for constructing meta-attack tasks and given the observation that different models focus on different discriminative regions in recognition \cite{dong2019evading}, the Perturbation Random Erasing (PRE) module is introduced to erase randomly selected perturbation regions to prevent the attacker from only learning to destroy the model-specific features or salient features, thus enhance the cross-model generalization of adversarial examples.
Meanwhile, the Normalization Mix (NorMix) strategy is devised to mimic cross-test embedding spaces by dynamically mixing the multi-domain batch-norm statistics of the target model, boosting attackers learning the ability of attacking target models that trained in different domain data.
Extensive experiments on numerous re-id benchmarks and models show our MTGA achieves state-of-the-art (SOTA) transferability on all six black-box attack scenarios,  demonstrating the effectiveness of our method. 
\textcolor{black}{Especially for cross-model\&dataset and cross-model\&dataset\&test attack, our MTGA surpasses the SOTA methods by 21.5\% and 11.3\% on mean mAP drop rate, respectively.}
In summary, our main contributions are as follows:
\begin{itemize}
  \item We propose a novel Meta Transferable Generative Attack (MTGA) method that creates extensive cross-model\&dataset\&test black-box meta attack tasks for adversarial generative attackers to learn to generate more generic and transferable adversarial examples against real-world re-id models.
  \item Cross-model\&dataset Attack Simulation approach is presented to mimic transfer-based cross-model and cross-dataset meta attack tasks by selecting distinct model and dataset for meta-train and meta-test processes.
  \item Perturbation Random Erasing module is devised to enhance the transferability by suppressing the model-specific features corruption and encouraging disruption of entire feature rather than only discriminative feature.
  \item Normalization Mix strategy is introduced to simulate cross-test attack by dynamically mixing the multi-domain batch-norm statistics of the target model, diversifying feature embedding spaces of re-id models. 
\end{itemize}

\section{Related Works}
\subsection{Transferable Adversarial Attack}
Szegedy \textit{et al.} \cite{szegedy2014intriguing} demonstrated the transferability of adversarial examples, enabling attackers to craft examples on surrogate models to attack target black-box models. Efforts to enhance adversarial transferability can be grouped into four categories: input transformation \cite{dong2019evading}, gradient modification \cite{dong2018boosting}, intermediate feature manipulation \cite{wang2021feature}, and model ensemble strategies \cite{xiong2022stochastic}. However, these methods focus solely on cross-model transferability, assuming consistent data distributions between attacked images and target model training data, which is rarely met in practical scenarios.
Cross-dataset transferability has received limited attention. Naseer \textit{et al.} \cite{naseer2019cross} proposed a generative network to produce cross-dataset perturbations by maximizing the fooling gap. Zhang \textit{et al.} \cite{zhang2021beyond} disrupted low-level features and improved transferability by randomly normalizing benign images. Li \textit{et al.} \cite{li2023cdta} employed self-supervised learning to train a domain-agnostic feature extractor for cross-dataset attacks. Yang \textit{et al.} \cite{yang2025prompt} leveraged vision-language models and prompt learning to enhance cross-dataset transferability. In contrast, our MTGA is designed for more complex cross-model\&dataset\&test attacks targeting black-box re-id models.

\subsection{Adversarial Attack Against Open-set Task}

Person re-id is a specialized image retrieval task focused on identifying a target individual across non-overlapping camera views \cite{cao2023event,xu2023deepchange,nguyen2024tackling,yang2023good,liu2024A,xu2022rank,yuan2023searching}. Unlike classification tasks, re-id operates as an open-set problem, where the test classes differ entirely from the training classes \cite{zheng2018open}. Previous attack methods \cite{moosavi2016deepfool,madry2018towards,carlini2017towards} on the image classification task are inapplicable to attack open-set task models \cite{zheng2023u}. To effectively attack open-set re-id, face recognition and image retrieval models, some white-box attack methods based on feature similarity \cite{bai2020adversarial, zheng2023u, bouniot2020vulnerability} and rank results disruption \cite{li2019universal,tolias2019targeted} have been developed. 
To accomplish black-box attacks against these models, researchers studied the transferable attacks on open-set tasks. Gong \textit{et al.}~\cite{gong2022person} used more obvious color variation to randomly disturb the retrieved images. Yang \textit{et al.}~\cite{yang2021learning} and Subramanyam ~\cite{subramanyam2023meta} enhanced the cross-dataset transferability by adopting multi-source datasets in additive and generative attack, respectively. Wang \textit{et al.}~\cite{wang2020transferable} developed a multi-stage discriminator network for cross-dataset general attack learning. Ding \textit{et al.}~\cite{ding2021beyond} introduced a model-insensitive regularization term for universal attack against different CNN structures. Yang \textit{et al.}~\cite{yang2022towards} built a combinatorial attack that consists of a functional color attack and universal additive attack to promote the cross-model\&dataset of the attack. Zhong \textit{et al.}~\cite{zhong2020towards} applied dropout layers to boost cross-model transferability. Li \textit{et al.}~\cite{li2023sibling} leveraged a highly related task as the sibling task to generate cross-model\&dataset transferable attacks. 
\textcolor{black}{Existing transfer-based open-set adversarial attack methods have incorporated considerations for diverse test data domains in open-set scenarios, along with cross-model architectural transferability. However, these approaches still fail to account for potential variations in the training domains of target models, which is particularly crucial for ensuring effectiveness in cross-test attack scenarios.}
\begin{figure*}[!t]
  \centering
  \includegraphics[width=1\linewidth]{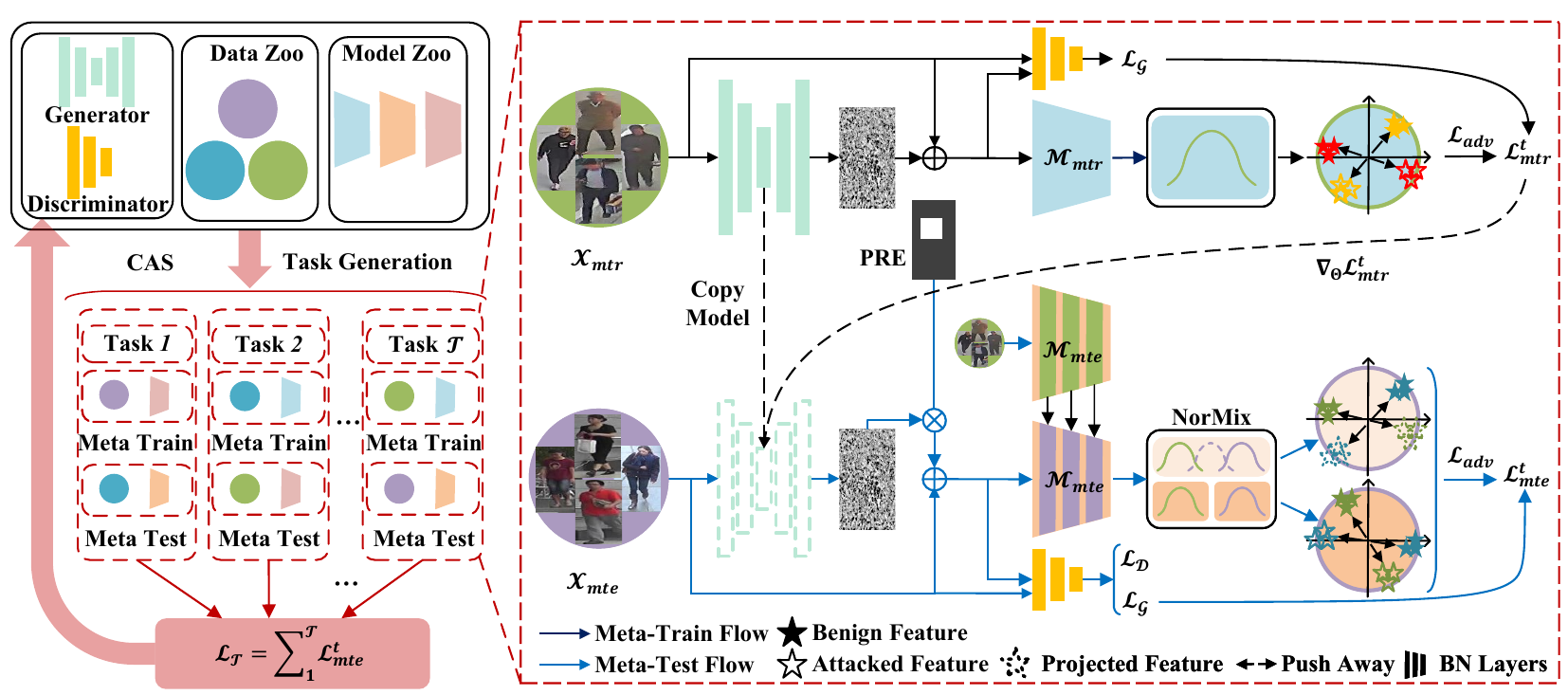}
  \caption{\textcolor{black}{The overall framework of our MTGA. CAS is applied to generate cross-model\&dataset meta attack tasks. In each task, the meta-train process calculates adversarial loss and generative loss as the meat-train loss and updates the copied generator by it. In meta-test process, Normalization Mix and Perturbation Random Erasing modules are conducted to promote the attacker possessing cross-test and cross-model transferability capability. The meta-test loss is calculated on the updated model and the sum of meta-test loss of all attack tasks are utilized to update the original adversarial generator.}}
  \label{fig:Fig2}
\end{figure*}
\subsection{Meta-learning}
Meta-learning is a learning-to-learn \cite{thrun1998learning} algorithm, which aims to improve further learning performance by distilling the experience from multiple learning episodes (\textit{i.e.}, meta-train and meta-test processes) \cite{hospedales2021meta,finn2017model}. It has been widely used in deep learning tasks, \textit{e.g.}, few-shot learning, domain generalization and hyperparameter optimization.
\textcolor{black}{Recently, some meta-learning based transferable adversarial attack methods have been proposed and show superiority to traditional attack method. Unlike traditional methods that train attacks on a single model or input, the meta-learning based method construct numerous meta transfer attack tasks using multiple models or inputs for training. They construct transferability error tests by meat-train and meta-test training tasks, instead of performing accuracy degradation attacks as in traditional methods, which makes the meta-learning based attack approaches to get better transferability.} 
Yuan \textit{et al.} \cite{yuan2021meta} enhanced the cross-model transferability by composing different cross-model meta attack tasks.
Fang \textit{et al.} \cite{fang2022learning} composed transfer attack tasks with data augmentation and model augmentation, through randomized data transformation and model backpropagation altering.
Yin \textit{et al.} \cite{yin2023generalizable} generalized the generic prior of examples by treating attack on each examples as one task and fine-tuning the surrogate model during the meta-test process.

\textcolor{black}{Distinct from above adversarial attack methods for open-set and meta-learning based attack methods, our method constructs extensive cross-model\&dataset\&test black-box adversarial attack tasks for attackers to learn how to generate more generic and transferable adversarial examples. And our CAS, PRE and NorMix modules are quite distinct from others.}
\label{subsec:overview}

\section{Methodology}
In this section, we first present the problem definition of the generative adversarial attack against re-id models in Section \ref{subsec:definition}. The overall framework of MTGA and the meta-learning optimization is then introduced in Section \ref{subsec:overview}. Right after that, the details about how to generate extensive transfer-based black-box meta attack tasks are described in Section \ref{subsec:generation}. Finally, the optimization procedure of our method are given in Section \ref{subsec:optimization}.

\subsection{Problem Definition}
\label{subsec:definition}

The goal of our proposed MTGA is to optimize the parameters $\bm{\theta}$ of the adversarial generator $\mathcal{G}$ to produce adversarial perturbation $\bm{\delta}$ for each benign image $\bm{x}$. The adversarial example $\bm{x^{adv}}$ is produced by adding additive perturbation to the query image to attack the re-id models $\mathcal{M}$ for outputting incorrect retrieval images. To ensure adversarial perturbations are imperceptible, the maximum magnitude of perturbations $\bm{\delta}$ allowed to be added cannot exceed $\epsilon$.
\begin{equation}
  \textcolor{black}{\bm{x^{adv}_{\theta}} = \mathcal{G}_{\bm{\theta}}(\bm{x})+\bm{x}, \quad\mathrm{s.t.}\|\bm{x^{adv}}-\bm{x}\|_\infty\leq\epsilon.}
  \label{eq:eq0}
\end{equation} 

The adversarial generator is first trained in the white-box way, knowing the attacked queries and the target re-id model. Then, it is fixed and used to produce perturbations for unseen data to attack black-box re-id models.

\subsection{Overall Framework}
\label{subsec:overview}

The proposed MTGA is based on the meta-learning optimization framework, as Fig. \ref{fig:Fig2} shows. Meta tasks $\mathcal{T}$ are generated to simulate the train-test processes of transfer-based black-box attack to train the generative attacker learning to produce generic adversarial examples. The data zoo $\mathcal{X}_z$ and model zoo $\mathcal{M}_z$ that contain multiple datasets and multiple re-id models are first prepared for meta-task generation. In each meta task $t$, datasets and re-id models for meta-train ($\mathcal{X}_{mtr}^t,\mathcal{M}_{mtr}^t$) and meta-test ($\mathcal{X}_{mte}^t,\mathcal{M}_{mte}^t$) processes are distinctly selected from the data zoo $\mathcal{X}_z$ and model zoo $\mathcal{M}_z$ to mimic training data and unseen test data, as well as the surrogate model and target model. The discriminator $\mathcal{D}$ is adopted in optimization processes to distinguish the adversarial images from benign images to boost generator $\mathcal{G}$ producing deceptive perturbations. The parameters $\bm{\theta}$ of generator $\mathcal{G}$ are updated after meta-train process.
Then, in the meta-test process, $\mathcal{G}$ generates adversarial perturbations for $\mathcal{X}_{mte}^t$ with the updated $\bm{\theta^{\prime}}$ to test the transferability of trained generator.
The perturbations are randomly erased by the PRE strategy and the features are projected to diverse embedding spaces through the NorMix module by mixing the $\mathcal{X}_{mtr}^t$ and $\mathcal{X}_{mte}^t$ feature distributions that extracted by $\mathcal{M}_{mte}^t$.
The meta-test errors of generated tasks serves as the training error of the various transfer-based black-box attack processes to optimize the adversarial generator.

\subsection{Meta Task Generation}
\label{subsec:generation}
The meta-task consists of a meta-training and a meta-testing process.
Meta-train process plays the role of transfer-based black-box attack training process, which utilizes white-box agent models and selected data to train the adversarial generator. And the meta-test process plays the role of transfer-based black-box attack testing process, which tests the transferability of the trained attacker against black-box target model and unseen images. By learning from generated black-box attack tasks, attackers can learn how to generate adversarial examples to attack black-box re-id models.
In terms of better learning for generating transferable and generalizable perturbations, a large number of meta-tasks that take all variations of realistic transfer-based black-box attacks into account should be constructed. Specifically, our approach generates diverse cross-model\&dataset\&test attack tasks by performing the following three methods.

\textbf{Cross-model\&dataset Attack Simulation method.}
Because of the unknown parameters of the re-id model and unseen domain queries to be attacked in black-box scenarios, the adversarial generator needs to learn to handle the cross-model and cross-dataset attack situations. To mimic this case, Cross-model\&dataset Attack Simulation method is proposed, which makes the target model and input data different during meta-train and meta-test process. Concretely, the data zoo and the model zoo that contains multiple datasets and multiple re-id models are constructed, from which CAS randomly selects distinct models and data for meta-train and meta-test processes to simulate cross-model and cross-dataset attacks. To  represent numerous models well, CAS takes baseline models of three mainstream approaches (\textit{i.e.}, global-based, part-based and attention-based) to construct the model zoo.

\textbf{Perturbation Random Erasing strategy.} Although there are several surrogate models in the model zoo to allow the attacker learning to handle cross-model attack scenarios, the number of these models is still limited, which may result in the attacker only learning to attack model-specific features. To address this problem, the Perturbation Random Erasing strategy is proposed. Base on the observation that different models tend to focus on distinct discriminative regions during recognition \cite{dong2019evading}, our PRE aims to prevent attacker from perturbing only model-specific feature regions by adding randomly erased incomplete perturbations on images, thereby boosting the attacker to disrupt holistic image features and enhancing the cross-model transferability of adversarial attacks. \textcolor{black}{Specifically, our PRE can be formulated by} 
\begin{equation}
  \textcolor{black}{\bm{x^{adv}_{\theta}} =  \bm{\mathcal{R}} \odot \mathcal{G}_{\bm{\theta}}(\bm{x})+\bm{x}, \quad\mathrm{s.t.}\|\bm{x^{adv}}-\bm{x}\|_\infty\leq\epsilon,}
  \label{eq:PRE}
\end{equation} 
\textcolor{black}{where $\bm{\mathcal{R}}$ is the random pattern and  $\odot$ is the Hadamard product. The random pattern $\bm{\mathcal{R}}$ is initially configured as a matrix of size $H/16 \times W/16$ with all elements set to 1, where $H$ and $W$ is the height and width of the person image $\bm{x}$. Then, with a probability $p$ for undergoing PRE, a randomly selected $m$ percentage of its elements are set to 0. Finally, the matrix is rescaled to the dimensions $H \times W$ using nearest-neighbor interpolation to get the patch masked random pattern $\bm{\mathcal{R}}$. By applying Hadamard product $\odot$ between the final random pattern and the perturbation, we can erase random patch regions of the perturbation. These incomplete perturbations prompt the attacker not to rely only on corrupting specific region features, as perturbations in these specific regions may be erased, leading to the failure of damaging specific region features.}

\textcolor{black}{
  With the PRE mechanisms, the gradient backpropagation for updating the adversarial generator parameters
  $\bm{\theta}$ with Eq.\ref{eq:eq4} $\mathcal{L}_{adv}$ loss can be formulated by:}
\textcolor{black}{
  \begin{equation}
    \resizebox{0.9\hsize}{!}{
      $\begin{aligned}
          \frac{\partial \mathcal{L}_{adv}}{\partial \bm{\theta}}
           & = \frac{\partial \mathcal{L}_{adv}}{\partial \bm{x^{adv}_\theta}} \cdot \frac{\partial \bm{x^{adv}_\theta}}{\partial \mathcal{G}_{\bm{\theta}}} \cdot \frac{\partial \mathcal{G}_{\bm{\theta}}}{\partial \bm{\theta}}                                                                                                                  \\
           & = \left( \frac{\partial \mathcal{L}_{adv}}{\partial \mathcal{M}(\bm{x^{adv}_\theta})} \cdot \frac{\partial \mathcal{M}(\bm{x^{adv}_\theta})}{\partial \bm{x^{adv}_\theta}} \right) \cdot \frac{\partial \bm{x^{adv}_\theta}}{\partial \mathcal{G}_{\bm{\theta}}} \cdot \frac{\partial \mathcal{G}_{\bm{\theta}}}{\partial \bm{\theta}} \\
           & = -\left( \frac{\partial \mathcal{E}}{\partial \mathcal{M}(\bm{x^{adv}_\theta})} \cdot \frac{\partial \mathcal{M}(\bm{x^{adv}_\theta})}{\partial \bm{x^{adv}_\theta}} \right) \cdot \bm{\mathcal{R}} \cdot \frac{\partial \mathcal{G}_{\bm{\theta}}}{\partial \bm{\theta}}                                                             \\
           & = -\bm{\mathcal{R}} \odot \left( \frac{\partial \mathcal{E}}{\partial \mathcal{M}(\bm{x^{adv}_\theta})} \cdot \frac{\partial \mathcal{M}(\bm{x^{adv}_\theta})}{\partial \bm{x^{adv}_\theta}} \right) \cdot \nabla_{\bm{\theta}} \mathcal{G}_{\bm{\theta}}(\bm{x})                                                                      \\
           & = -\bm{\mathcal{R}} \odot \left( \left( \mathcal{M}(\bm{x}) - \mathcal{M}(\bm{x^{adv}_\theta}) \right) \cdot \frac{\partial \mathcal{M}(\bm{x^{adv}_\theta})}{\partial \bm{x^{adv}_\theta}} \right) \cdot \nabla_{\bm{\theta}} \mathcal{G}_{\bm{\theta}}(\bm{x}).
        \end{aligned}$
    }
    \label{eq:grad_chain}
  \end{equation}
  It can be observed that our PRE, by introducing $\mathcal{R}$, enables the stochastic sparsification of gradients, thereby suppressing the learning on model-specific salient features and facilitating the disruption of comprehensive features of the agent model $\mathcal{M}$. From another perspective, the perturbations after random erasure can be regarded as perturbations generated by different generators. This is equivalent to training multiple generators, ultimately resulting in an implicit ensemble-averaged output, which reduces the variance of the attack effectiveness and renders the generated perturbations more generalizable. }

\textcolor{black}{PRE is adopted in the meta-test process to test the attack error of trained adversarial attackers with generated incomplete perturbations, optimizing that error will enhance the attacker to achieve holistic destruction of image features and improve the transferability against black-box models. }

\textbf{Normalization Mix module.} The models that trained with different domain data could project person images to various feature embeddings, even though they share the same model architecture. NorMix is devised to project features to different feature embedding spaces, which is applied in meta-test process to promote the attacker learning to handle this cross-test issue. 
NorMix is motivated by the insight that the weight matrix typically captures label information, whereas the Batch Normalization (BN) layer \cite{ioffe2015batch} houses domain-specific knowledge \cite{li2016revisiting}. Building on this, we introduce the Normalization Mix module to simulate various feature embeddings by blending the BN statistics, which reflecting the underlying distribution of the model's training data. 

\textcolor{black}{Specifically, there are multiple batch-norm layers across the re-id model architectures, and the batch normalization is formulated as}
\begin{equation}
  \textcolor{black}{\bm{\hat{f}}=\bm{\gamma}\frac{\bm{f}-\bm{\mu}}{\bm{\sigma}}+\bm{\beta},}
  \label{eq:1}
\end{equation}
\textcolor{black}{where $\bm{f}$ is the input feature, $\bm{\mu}$ and $\bm{\sigma}$ are the mean and variance of $\bm{f}$, $\bm{\gamma$} and $\bm{\beta}$ are learnable affine parameters used for linear transformation, and $\bm{\hat{f}}$ is the output feature after batch normalization. Once the model finishes training, the BN statistics remain unchanged and the model maps input data to a consistent feature space. To get diverse feature embeddings that the test data may be projected by the target model, the statistic of each batch-norm layer is mixed by}
\begin{equation}
  \textcolor{black}{\bm{\sigma_{mix}}=\lambda\bm{\sigma_{mte}}+(1-\lambda)\bm{\sigma_{mtr}},}
  \label{eq:2}
\end{equation}
\begin{equation}
  \textcolor{black}{\bm{\mu_{mix}}=\lambda\bm{\mu_{mte}}+(1-\lambda)\bm{\mu_{mtr}},}
  \label{eq:3}
\end{equation}
\textcolor{black}{where $\bm{\mu_{mte}}$ and $\bm{\sigma_{mte}}$ are the empirical mean and variance of the pretrained meta-test model $\mathcal{M}_{mte}$, $\bm{\mu_{mtr}}$ and $\bm{\sigma_{mtr}}$ are the training statistics of the meta-train datasets $\mathcal{X}_{mtr}$ on $\mathcal{M}_{mte}$ and $\lambda$ is the mix coefficient that sampled from Beta Distribution. Because the training dataset of the $\mathcal{M}_{mte}$ and the meta-train dataset $\mathcal{X}_{mtr}$ are different, we can get effective mixed BN statistics by mixing them. With the dynamical mix coefficient $\lambda$ and different meta-train datasets $\mathcal{X}_{mtr}$, diverse mixed mean $\bm{\mu_{mix}}$ and variance $\bm{\sigma_{mix}}$ can be obtained. Finally, meta-test data features $\bm{f_{mte}}$ can be embedded to different feature spaces by}
\begin{equation}
  \textcolor{black}{\bm{\hat{f_{mte}}}=\bm{\gamma_{mte}}\frac{\bm{f_{mte}}-\bm{\mu_{mix}}}{\bm{\sigma_{mix}}}+\bm{\beta_{mte}},}
  \label{eq:4}
\end{equation}
\textcolor{black}{where $\bm{\gamma_{mte}}$ and $\bm{\beta_{mte}}$ are copied from the batch-norm layers of meta-test model, $\bm{\hat{f_{mte}}}$ is the output features mapped to diverse embeddings. By leveraging $\bm{\hat{f_{mte}}}$ derived from various feature spaces throughout meta-test process, our MTGA can facilitate the adversarial generators learning cross-test transfer capability.}

\textcolor{black}{NorMix can also be viewed as a generalization of test-time adaptation (TTA) methods \cite{kang2024membn,limttn}, which aim to adapt models to unseen test domains in real-time by dynamically adjusting normalization statistics while preserving the original network parameters. TTA approaches address test and training data distribution shift by replacing the original BN parameters with statistics estimated from test batches. NorMix aligns with TTA's principle that domain-specific knowledge resides primarily in BN layers, adapting models' feature embeddings to diverse distributions by dynamically adjusting BN statistics while preserving network weights, thereby validating that normalization-layer adjustments can simulate cross-test feature distributions.} 

\begin{algorithm}[t]
  \small
  \caption{Meta Transferable Generative Attack algorithm}
  \label{alg:algorithm}
  \renewcommand{\algorithmicrequire}{\textbf{Input:}}
  \renewcommand{\algorithmicensure}{\textbf{Output:}}
  \begin{algorithmic}[1] \color{black}%[1] enables lSine numbers
    \REQUIRE Data zoo $\mathcal{X}_z$, model zoo $\mathcal{M}_z$, generator $\mathcal{G}$, discriminator $\mathcal{D}$
    \ENSURE  Generative adversarial attacker $\mathcal{G}$
    \STATE Initialize parameters $\bm{\theta}$ of $\mathcal{G}$, $\bm{\varphi}$ of $\mathcal{D}$, learning rate $\eta$ of inner loop, $\alpha$ of outer loop
    \FOR{$i$=0 to $\mathcal{I}$-$1$}
    \FOR{$t$ = 0 to $\mathcal{T}$-1}
    % \STATE \textit{$\%$meta-Task Generation}
    \STATE Sample two models $\mathcal{M}_{mtr}, \mathcal{M}_{mte}$ and two batch data $\mathcal{X}_{mtr}, \mathcal{X}_{mte}$ from $\mathcal{M}_z$ and $\mathcal{X}_z$ 
    \STATE \textit{$\%$Meta-train}
    \STATE Calculate meta-train loss $\scalebox{0.96}{$\mathcal{L}_{mtr}^{t}(\bm{\theta},\bm{\varphi},\mathcal{X}_{mtr}^t,\mathcal{M}_{mtr})$}$ by Eq.\ref{eq:eq5}
    % \STATE \textit{$\%$Meta-train Task Generation}
    \STATE Update parameters $\scalebox{0.96}{$\bm{\theta^{\prime}}=\bm{\theta}-\eta\bm{\nabla_\theta} \mathcal{L}_{mtr}^t$}$
    \STATE \textit{$\%$Meta-test}
    \STATE Do Perturbation Random Erasing and Normalization Mix
    \STATE Calculate meta-test loss $\scalebox{0.95}{$\mathcal{L}_{mte}^{t}(\bm{\theta^{\prime}},\bm{\varphi},\mathcal{X}_{mte}^t,\mathcal{M}_{mte})$}$ by Eq.\ref{eq:eq7}
    \STATE Calculate discrimination loss $\scalebox{0.96}{$\mathcal{L}_\mathcal{D}^{t}(\bm{\theta^{\prime}},\bm{\varphi},\mathcal{X}_{mte}^t)$}$ by Eq.\ref{eq:eq3}
    \ENDFOR
    \STATE Update parameters $\scalebox{0.96}{$\bm{\theta}\leftarrow\bm{\theta}-\alpha\bm{\nabla_{\theta}}\frac1{\mathcal{T}}\sum\nolimits_{1}^{\mathcal{T}} \mathcal{L}_{mte}^{t}$}$
    \STATE Update parameters $\scalebox{0.96}{$\bm{\varphi}\leftarrow\bm{\varphi}-\alpha\bm{\nabla_\varphi} \frac1{\mathcal{T}}\sum\nolimits_{1}^{\mathcal{T}} \mathcal{L}_\mathcal{D}^{t}$}$
    \ENDFOR
    % \ENDWHILE
  \end{algorithmic}
\end{algorithm}
\subsection{Optimization Procedure}
\label{subsec:optimization}
The parameters $\bm{\theta}$ of adversarial generator $\mathcal{G}$ are supposed to be optimized by the meta-learning optimization. To disrupt the retrieval list of generated adversarial examples, the attacked image features should be far away from the original features. 
\textcolor{black}{In our MTGA, the adversarial Euclidean Distance loss}
\begin{equation}
  \textcolor{black}{\mathcal{L}_{adv}(\bm{\theta},\mathcal{M},\bm{x}) = -\mathcal{E}(\mathcal{M}(\bm{x^{adv}_\theta}),\mathcal{M}(\bm{x})),}
  \label{eq:eq4}
\end{equation}
\textcolor{black}{is applied to corrupt the similarity of adversarial features $\mathcal{M}(\bm{x^{adv}_\theta})$ and benign features $\mathcal{M}(\bm{x})$ extracted by the re-id model $\mathcal{M}$, where $\mathcal{E}$ is the Euclidean distance.}
Meanwhile, $\mathcal{G}$ and $\mathcal{D}$ are trained by the GAN loss respectively, denote as:
\begin{equation}
  \textcolor{black}{\mathcal{L}_\mathcal{G}(\bm{\theta},\bm{\varphi},\bm{x}) = \log(1-\mathcal{D}_{\bm{\varphi}}(\bm{x^{adv}_{\theta}}),}
  \label{eq:eq2}
\end{equation}
\begin{equation}
  \textcolor{black}{\mathcal{L}_\mathcal{D}(\bm{\theta},\bm{\varphi},\bm{x}) = \log \mathcal{D}_{\bm{\varphi}}(\bm{x})+\log(1-\mathcal{D}_{\bm{\varphi}}(\bm{x^{adv}_{\theta}})).}
  \label{eq:eq3}
\end{equation}
\textbf{Meta-train.} With the $\mathcal{X}_{mtr}$ and $\mathcal{M}_{mtr}$, the objective function of meta-train process is calculated by
\begin{equation}
  \textcolor{black}{\mathcal{L}_{mtr}^t = \mathcal{L}_\mathcal{G}^t(\bm{\theta},\bm{\varphi},\mathcal{X}_{mtr}^t)+ \mathcal{L}_{adv}^t(\bm{\theta},\mathcal{M}_{mtr}^t,\mathcal{X}_{mtr}^t).}
  \label{eq:eq5}
\end{equation}
\textbf{Meta-test.} After meta-train process, the parameters $\bm{\theta}$ of $\mathcal{G}$ is updated to $\bm{\theta^\prime}$, and meta-test loss is expressed by
\begin{equation}
  \textcolor{black}{\mathcal{L}_{mte}^t = \mathcal{L}_\mathcal{G}^t(\bm{\theta^{\prime}},\bm{\varphi},\mathcal{X}_{mte}^t)+ \mathcal{L}_{adv}^t(\bm{\theta^{\prime}},\mathcal{M}_{mte}^t,\mathcal{X}_{mte}^t).}
  \label{eq:eq7}
\end{equation}
\textbf{Meta Optimization.} The final loss consists of the meta-test errors for each meta-task,  formulated as
\begin{equation}
  \textcolor{black}{\mathcal{L}_{\bm{\theta}} = \frac1{\mathcal{T}}\sum\nolimits_{t=1}^{\mathcal{T}}\mathcal{L}_{mte}^{t}
  \label{eq:eq8},}
\end{equation}
which represents the error of adversarial generator with parameters $\bm{\theta}$ for different cases of transfer-based black-box attacks. By optimizing the $\mathcal{L}_{\bm{\theta}}$, adversarial generator that produces highly transferable adversarial examples against different black-box re-id models can be learned. The optimization procedure is summarized in Algorithm \ref{alg:algorithm}.

\begin{table*}[t]
  \caption{Six black-box attack settings in our experiments. The \ding{52} and \ding{56} signs for query domain, model architecture and model domain represent whether these black-box test settings are the same as the corresponding settings in the white-box training process. The \ding{52} and \ding{56} signs for test-domain indicate whether the domain of the query images and the domain of the model training data are consistent during black-box attacking. The implement details of these settings in our experiments are shown in the right half of table, where $\mathcal{M}_b$(Market) represents the black-box re-id models that trained on Market dataset. The arch and Duke are abbreviations for architecture and DukeMTMC.}
  \centering
  % \resizebox{1.0\textwidth}{!}{

    \begin{tabular}{c|cccc|cccc}
      \hline
      Attack Settings            & \begin{tabular}{c}
                                     Query \\
                                     domain
                                   \end{tabular} & \begin{tabular}{c}
                                                     Model \\
                                                     arch
                                                   \end{tabular} & \begin{tabular}{c}
                                                                     Model \\
                                                                     domain
                                                                   \end{tabular} & \begin{tabular}{c}
                                                                                     Test \\
                                                                                     domain
                                                                                   \end{tabular} & \begin{tabular}{c}
                                                                                                     Training \\
                                                                                                     data
                                                                                                   \end{tabular} & \begin{tabular}{c}
                                                                                                                     Surrogate \\
                                                                                                                     model
                                                                                                                   \end{tabular}    & \begin{tabular}{c}
                                                                                                                                        Target \\
                                                                                                                                        data
                                                                                                                                      \end{tabular} & \begin{tabular}{c}
                                                                                                                                                        Target \\
                                                                                                                                                        model
                                                                                                                                                      \end{tabular}                                          \\

      \hline
      % Source attack & \ding{52} & \ding{52} & \ding{52} & \ding{52} \\
      Cross-dataset              & \ding{56}       & \ding{52}         & \ding{56}       & \ding{52}         & $\mathcal{X}_z$    & $\mathcal{M}_z$(Duke) & Market             & $\mathcal{M}_z$(Market) \\
      Cross-dataset\&test        & \ding{56}       & \ding{52}         & \ding{56}       & \ding{56}       & $\mathcal{X}_z$    & $\mathcal{M}_z$(Duke) & VIPeR              & $\mathcal{M}_z$(Market) \\
      Cross-model                & \ding{52}         & \ding{56}       & \ding{52}         & \ding{52}         & $\mathcal{X}_z$    & $\mathcal{M}_z$(Duke) & Duke               & $\mathcal{M}_b$(Duke)   \\
      Cross-model\&test          & \ding{52}         & \ding{56}       & \ding{56}       & \ding{56}       & $\mathcal{X}_z$    & $\mathcal{M}_z$(Duke) & Duke               & $\mathcal{M}_b$(Market) \\
      Cross-model\&dataset       & \ding{56}       & \ding{56}       & \ding{56}       & \ding{52}         & $\mathcal{X}_z$    & $\mathcal{M}_z$(Duke) & Market             & $\mathcal{M}_b$(Market) \\
      Cross-model\&dataset\&test & \ding{56}       & \ding{56}       & \ding{56}       & \ding{56}       & $\mathcal{X}_z$    & $\mathcal{M}_z$(Duke) & VIPeR              & $\mathcal{M}_b$(Market) \\
      \hline
    \end{tabular}
  % }
  \label{tb:tb1}
\end{table*}

\section{Experiments}
\textcolor{black}{To evaluate the superiority of our method, we first provide training and evaluation settings in the Section \ref{subsec:Experimental Setup} and then present experimental results in Section \ref{subsec:Experimental Results}. Afterwards, comprehensive evaluations including ablation studies, adversarial example quality assessment, visualization analyses, and attack effectiveness against defense mechanisms are provided to further validate the efficacy of our proposed method.}  

\subsection{\textcolor{black}{Experimental Setup}}
\label{subsec:Experimental Setup}
\textcolor{black}{\textbf{Training details.}} Model zoo is composed of IDE \cite{zheng2016person}, PCB \cite{sun2018beyond} and ViT \cite{he2021transreid}, which are all trained on the DukeMTMC \cite{ristani2016performance} datasets. And the data zoo consists of DukeMTMC \cite{ristani2016performance}, CUHK03 \cite{li2014deepreid}, and MSMT17 \cite{Wei2018Person} datasets.
MAML \cite{finn2017model} is adopted as our meta-learning framework and in each iteration $5$ meta-tasks are generated. Adam \cite{kingma2014adam} optimizer is employed to optimize the model parameters. The learning rate of inner loop $\eta$ and outer loop $\alpha$ are set to 1e-4 and 2e-4. The generator and discriminator model are referenced to the Mis-Ranking \cite{wang2020transferable}. All experiments are performed by $\mathcal{L}_{\infty}$-bounded attacks with $\epsilon=8/255$, where $\epsilon$ is the upper bound for the change of each pixel. The mix coefficient of NorMix is sampled from Beta Distribution, \textit{i.e.}, $\lambda\sim\mathrm{Beta}(5,5)$. \textcolor{black}{The probability $p$ of undergoing PRE is set to $0.8$ and the mask percentage $m$ of random pattern is set to $0.2$.}

\textcolor{black}{\textbf{Evaluation settings.}} To verify the attack performance of our methods against real-world re-id models, we comprehensively consider different adversarial attack scenarios and set up six attack settings. The details of these settings are showed in Tab. \ref{tb:tb1}.
The cross-model attack setting implies the black-box target model architecture is different with the surrogate model, yet the training domain of them is the same. The cross-dataset attack setting means the domain of query images and re-id models are different from the white-box attack training process, and query images and the target model training data are in the same domain. These settings are the same as transfer-based black-box re-id attacks proposed by \cite{yang2022towards}, to which we have added cross-test setting. The cross-test setting indicates that the domains of the query data and the target model are different, simulating the most practical application of the real-world re-id models.

\textcolor{black}{\textbf{Evaluation models and datasets.}}
To evaluate the transferability of our adversarial generator to different re-id models, numerous re-id models $\mathcal{M}_B$ (\textit{i.e.}, BOT \cite{luo2019bag}, LSRO \cite{zheng2017unlabeled}, MuDeep \cite{qian2017multi}, Aligned \cite{zhang2017alignedreid}, MGN \cite{wang2018learning}, HACNN \cite{li2018harmonious}, Transreid \cite{he2021transreid}, PAT \cite{ni2023part}) are taken to act as the black-box re-id models. Notably, \textbf{these models are in different backbones}, including ResNet \cite{he2016deep} (\textit{i.e.}, BOT \cite{luo2019bag}), ViT \cite{dosovitskiy2020image} (\textit{i.e.}, Transreid \cite{he2021transreid}, PAT \cite{ni2023part}), DenseNet \cite{huang2017Densely} (\textit{i.e.}, LSRO \cite{zheng2017unlabeled}) and Inception-v3 \cite{szegedy2016rethinking} (\textit{i.e.}, MuDeep \cite{qian2017multi}). Also, \textbf{these models are in different architectures}, including global-based (\textit{i.e.}, BOT \cite{luo2019bag}), part-based (\textit{i.e.}, MGN \cite{wang2018learning}) and attention-based (\textit{i.e.}, HACNN \cite{li2018harmonious}). In order to test the transferabilities on different domain models, \textbf{these models are trained on different domain datasets} (\textit{i.e.}, Market \cite{zheng2015scalable} and DukeMTMC \cite{ristani2016performance}). Meanwhile, to test the transferability of our attacker to unseen queries, VIPeR \cite{gray2007evaluating} and  Market \cite{zheng2015scalable} datasets play the role of unseen domain data.

\begin{table*}[t]
  \begin{minipage}[t]{0.49\textwidth}
    \vspace{0.20cm}
    \centering
    \caption{Results of \textbf{cross-dataset} attack. The best performance is in \textbf{bold}.}
    \label{tb:tb2}
    % \resizebox{1\textwidth}{!}{
      \begin{tabular}{c|ccc|cc}
        \hline
        Methods & IDE & PCB & ViT & aAP$\downarrow$ & mDR$\uparrow$ \\
        \hline
        None & 75.5 & 70.7 & 86.5 & 77.6 & - \\
        \hline
        \textcolor{black}{GAP} & \textcolor{black}{10.4} & \textcolor{black}{-} & \textcolor{black}{-} & \textcolor{black}{-} & \textcolor{black}{-} \\
        \textcolor{black}{CDA} & \textcolor{black}{13.3} & \textcolor{black}{-} & \textcolor{black}{-} & \textcolor{black}{-} & \textcolor{black}{-} \\
        \textcolor{black}{LTP} & \textcolor{black}{9.1} & \textcolor{black}{-} & \textcolor{black}{-} & \textcolor{black}{-} & \textcolor{black}{-} \\
        \textcolor{black}{BIA} & \textcolor{black}{14.8} & \textcolor{black}{-} & \textcolor{black}{-} & \textcolor{black}{-} & \textcolor{black}{-} \\
        \textcolor{black}{PDCL-Attack} & \textcolor{black}{7.4} & \textcolor{black}{-} & \textcolor{black}{-} & \textcolor{black}{-} & \textcolor{black}{-} \\
        \hline
        MetaAttack & 4.2 & - & - & - & - \\
        Mis-Ranking & 26.9 & - & - & - & - \\
        MUAP & 19.3 & - & - & - & - \\
        \hline
        MetaAttack* & 20.2 & 35.8 & 61.1 & 39.0 & 49.7 \\
        Mis-Ranking* & 16.8 & 36.8 & 48.4 & 34.0 & 56.1 \\
        MUAP* & 14.0 & 26.0 & 42.1 & 27.4 & 64.7 \\
        \hline
        MTGA* & 17.1 & 26.6 & 43.7 & 29.1 & 62.5 \\
        MTGA(Ours) & \textbf{10.8} & \textbf{25.5} & \textbf{38.4} & \textbf{24.9} & \textbf{67.9} \\
        \hline
    \end{tabular}
    % }

  \end{minipage}
  % \hfill
  \begin{minipage}[t]{0.489\textwidth}
    \centering
    \vspace{0.20cm}

    \caption{Results of \textbf{cross-dataset\&test} attack. The best performance is in \textbf{bold}.}
    \label{tb:tb3}
    % \resizebox{1.0\textwidth}{!}{
      \begin{tabular}{c|ccc|cc}
        \hline
        Methods & IDE & PCB & ViT & aAP$\downarrow$ & mDR$\uparrow$ \\
        \hline
        None & 30.0 & 33.0 & 51.0 & 38.0 & - \\
        \hline
        \textcolor{black}{GAP} & \textcolor{black}{12.7} & \textcolor{black}{-} & \textcolor{black}{-} & \textcolor{black}{-} & \textcolor{black}{-} \\
        \textcolor{black}{CDA} & \textcolor{black}{12.6} & \textcolor{black}{-} & \textcolor{black}{-} & \textcolor{black}{-} & \textcolor{black}{-} \\
        \textcolor{black}{LTP} & \textcolor{black}{9.9} & \textcolor{black}{-} & \textcolor{black}{-} & \textcolor{black}{-} & \textcolor{black}{-} \\
        \textcolor{black}{BIA} & \textcolor{black}{12.3} & \textcolor{black}{-} & \textcolor{black}{-} & \textcolor{black}{-} & \textcolor{black}{-} \\
        \textcolor{black}{PDCL-Attack} & \textcolor{black}{11.1} & \textcolor{black}{-} & \textcolor{black}{-} & \textcolor{black}{-} & \textcolor{black}{-} \\
        \hline
        MetaAttack & 10.0 & - & - & - & - \\
        Mis-Ranking & 14.2 & - & - & - & - \\
        MUAP & 11.9 & - & - & - & - \\
        \hline
        MetaAttack* & 14.1 & 24.7 & 40.7 & 26.5 & 30.3 \\
        Mis-Ranking* & 12.4 & 25.9 & 34.4 & 24.2 & 36.2 \\
        MUAP* & 11.9 & 20.4 & 35.9 & 22.7 & 40.2 \\
        \hline
        MTGA* & 12.7 & 22.4 & 33.0 & 22.7 & 40.3 \\
        MTGA(Ours) & \textbf{10.4} & \textbf{21.9} & \textbf{30.7} & \textbf{21.0} & \textbf{44.7} \\
        \hline
    \end{tabular}
    % }
  \end{minipage}
\end{table*}

\textcolor{black}{\textbf{Evaluation metrics.}}
The adversarial attack performance of the generated adversarial samples against different re-id models is measured by three metrics, mean Average Precision (mAP) \cite{zheng2015scalable}, average mAP (aAP) and mean mAP Drop Rate (mDR) \cite{ding2021beyond}. The aAP is calculated by
\begin{equation}
  aAP =\ \frac{\sum_{i=0}^{N}{mAP}_i}{N},
\end{equation}
where ${mAP}_i$ represents mAP of the $i$-th re-id models. The mDR is designed to show the success rate of the adversarial attacks to multiple re-id models and is formulated as
\begin{equation}
  mDR =\frac{aAP-aAP_{adv}}{aAP},
\end{equation}
where $aAP$ is the aAP of the re-id models on the benign images and $aAP_{adv}$ is on the generated adversarial examples.
\textcolor{black}{Smaller aAP and larger mDR represent better transferability of adversarial examples, so we use aAP$\downarrow$ and mDR$\uparrow$ to indicate this relationship in the table of experimental results for clearer comparisons.}

\begin{table*}[t]
  \centering
  \vspace{0.15cm}

  \caption{Results of \textbf{cross-model} attack. The best performance is in \textbf{bold}.}
  \label{tb:tb4}
  % \resizebox{1\textwidth}{!}{
    \begin{tabular}{c|ccc|cc|ccc|cc}
      \hline
      \multirow{2}{*}{ Methods } & \multicolumn{3}{c|}{ Global-based } & \multicolumn{2}{c|}{ Part-based } & \multicolumn{3}{c|}{ Attention-based } & \multirow{2}{*}{ aAP$\downarrow$  } & \multirow{2}{*}{ mDR$\uparrow$ } \\
      \cline{2-9} & BOT & LSRO & MuDeep & Aligned & MGN & HACNN & Transreid & PAT & & \\
      \hline
      None & 76.2 & 55.0 & 43.0 & 69.7 & 66.2 & 60.2 & 79.6 & 70.6 & 65.0 & - \\
      \hline
      \textcolor{black}{GAP} & \textcolor{black}{12.9} & \textcolor{black}{14.6} & \textcolor{black}{13.7} & \textcolor{black}{24.5} & \textcolor{black}{16.4} & \textcolor{black}{16.5} & \textcolor{black}{46.7} & \textcolor{black}{45.8} & \textcolor{black}{23.9} & \textcolor{black}{63.3} \\
      \textcolor{black}{CDA} & \textcolor{black}{9.6} & \textcolor{black}{12.5} & \textcolor{black}{12.7} & \textcolor{black}{20.8} & \textcolor{black}{14.7} & \textcolor{black}{15.0} & \textcolor{black}{42.3} & \textcolor{black}{40.8} & \textcolor{black}{21.1} & \textcolor{black}{67.6} \\
      \textcolor{black}{BIA} & \textcolor{black}{14.3} & \textcolor{black}{33.1} & \textcolor{black}{24.5} & \textcolor{black}{44.9} & \textcolor{black}{58.0} & \textcolor{black}{41.9} & \textcolor{black}{71.3} & \textcolor{black}{60.8} & \textcolor{black}{43.6} & \textcolor{black}{32.9} \\
      \textcolor{black}{LTP} & \textcolor{black}{12.3} & \textcolor{black}{22.3} & \textcolor{black}{23.3} & \textcolor{black}{30.9} & \textcolor{black}{37.8} & \textcolor{black}{22.5} & \textcolor{black}{49.6} & \textcolor{black}{45.5} & \textcolor{black}{30.5} & \textcolor{black}{53.0} \\
      \textcolor{black}{PDCL-Attack} & \textcolor{black}{11.8} & \textcolor{black}{11.1} & \textcolor{black}{10.5} & \textcolor{black}{22.3} & \textcolor{black}{12.6} & \textcolor{black}{14.2} & \textcolor{black}{37.5} & \textcolor{black}{32.0} & \textcolor{black}{19.0} & \textcolor{black}{70.8} \\
      \hline
      MetaAttack & 14.9 & 44.0 & 31.8 & 49.5 & 57.4 & 54.6 & 75.3 & 64.5 & 49.0 & 24.6 \\
      Mis-Ranking & 14.4 & 6.8 & 8.0 & 16.5 & 8.4 & 8.8 & 34.5 & 42.9 & 17.5 & 73.1 \\
      MUAP & 16.3 & 9.2 & 11.1 & 23.1 & 11.4 & 13.8 & 34.2 & 40.4 & 19.9 & 69.4 \\
      \hline
      MetaAttack* & 23.2 & 15.0 & 11.7 & 22.9 & 13.6 & 19.6 & 43.6 & 40.8 & 23.8 & 63.4 \\
      Mis-Ranking* & 6.8 & 2.0 & 9.9 & 8.7 & 4.3 & 6.6 & 16.3 & 22.3 & 9.6 & 85.2 \\
      MUAP* & 18.6 & 8.2 & 8.5 & 16.5 & 7.0 & 11.4 & 29.9 & 32.0 & 16.5 & 74.6 \\
      \hline
      MTGA* & 7.9 & 3.1 & 7.8 & 8.7 & 4.4 & 4.9 & 15.0 & 23.2 & 9.4 & 85.5 \\
      MTGA(Ours) & \textbf{5.1} & \textbf{1.4} & \textbf{7.2} & \textbf{6.5} & \textbf{3.2} & \textbf{4.9} & \textbf{13.8} & \textbf{19.9} & \textbf{7.7} & \textbf{88.2} \\
      \hline
  \end{tabular}
  % }
\end{table*}

\begin{table*}[t]
  \centering
  \vspace{0.15cm}

  \caption{Results of \textbf{cross-model\&test} attack. The best performance is in \textbf{bold}.}
  \label{tb:tb5}
  % \resizebox{1\textwidth}{!}{
    \begin{tabular}{c|ccc|cc|ccc|cc}
      \hline
      \multirow{2}{*}{ Methods } & \multicolumn{3}{c|}{ Global-based } & \multicolumn{2}{c|}{ Part-based } & \multicolumn{3}{c|}{ Attention-based } & \multirow{2}{*}{ aAP$\downarrow$ } & \multirow{2}{*}{ mDR$\uparrow$ } \\
      \cline{2-9} & BOT & LSRO & MuDeep & Aligned & MGN & HACNN & Transreid & PAT & & \\
      \hline
      None & 14.9 & 13.5 & 4.5 & 18.3 & 22.3 & 11.2 & 43.6 & 44.6 & 21.6 & - \\
      \hline
      \textcolor{black}{GAP} & \textcolor{black}{7.2} & \textcolor{black}{8.1} & \textcolor{black}{2.8} & \textcolor{black}{11.4} & \textcolor{black}{13.2} & \textcolor{black}{6.4} & \textcolor{black}{26.9} & \textcolor{black}{29.9} & \textcolor{black}{13.2} & \textcolor{black}{38.7} \\
      \textcolor{black}{CDA} & \textcolor{black}{7.7} & \textcolor{black}{8.5} & \textcolor{black}{2.9} & \textcolor{black}{11.7} & \textcolor{black}{13.7} & \textcolor{black}{7.0} & \textcolor{black}{27.1} & \textcolor{black}{30.2} & \textcolor{black}{13.6} & \textcolor{black}{37.0} \\
      \textcolor{black}{BIA} & \textcolor{black}{7.9} & \textcolor{black}{8.8} & \textcolor{black}{3.7} & \textcolor{black}{11.9} & \textcolor{black}{17.6} & \textcolor{black}{9.2} & \textcolor{black}{34.3} & \textcolor{black}{37.1} & \textcolor{black}{16.3} & \textcolor{black}{24.5} \\
      \textcolor{black}{LTP} & \textcolor{black}{6.6} & \textcolor{black}{8.9} & \textcolor{black}{3.3} & \textcolor{black}{12.0} & \textcolor{black}{15.4} & \textcolor{black}{7.5} & \textcolor{black}{27.2} & \textcolor{black}{29.3} & \textcolor{black}{13.8} & \textcolor{black}{36.2} \\
      \textcolor{black}{PDCL-Attack} & \textcolor{black}{\textbf{4.8}} & \textcolor{black}{5.3} & \textcolor{black}{2.6} & \textcolor{black}{8.3} & \textcolor{black}{10.4} & \textcolor{black}{5.0} & \textcolor{black}{24.5} & \textcolor{black}{25.2} & \textcolor{black}{10.8} & \textcolor{black}{50.2} \\
      \hline
      MetaAttack & 4.9 & 11.8 & 4.3 & 12.6 & 19.9 & 10.8 & 41.3 & 40.1 & 18.2 & 15.7 \\
      Mis-Ranking & 9.2 & 6.4 & 2.1 & 9.9 & 11.3 & 5.5 & 29.3 & 35.4 & 13.6 & 37.0 \\
      MUAP & 7.2 & 5.9 & 2.6 & 10.4 & 10.4 & 6.0 & 28.4 & 31.9 & 12.9 & 40.3 \\
      \hline
      MetaAttack* & 6.5 & 5.5 & 2.9 & 8.7 & 10.1 & 6.4 & 30.1 & 31.2 & 12.6 & 41.7 \\
      Mis-Ranking* & 6.7 & 4.5 & 2.3 & 8.3 & 7.9 & 4.0 & 22.0 & 26.5 & 10.3 & 52.3 \\
      MUAP* & 5.0 & 3.5 & 2.3 & 8.5 & 7.5 & 4.8 & 22.5 & 24.7 & 9.9 & 54.4 \\
      \hline
      MTGA* & 6.7 & 4.8 & 1.9 & 7.5 & 7.6 & 3.4 & 20.6 & 25.4 & 9.7 & 55.1 \\
      MTGA(Ours) & \textbf{5.5} & \textbf{3.4} & \textbf{1.9} & \textbf{7.0} & \textbf{6.3} & \textbf{3.4} & \textbf{18.7} & \textbf{23.6} & \textbf{8.7} & \textbf{59.7} \\
      \hline
  \end{tabular}
  
  % }
\end{table*}
\subsection{\textcolor{black}{Experimental Results}}
\label{subsec:Experimental Results}

\textcolor{black}{We compare our proposed MTGA method with state-of-the-art attack methods on transferable black-box  re-id attacks, including MUAP \cite{ding2021beyond}, Mis-Ranking \cite{wang2020transferable}, MetaAttack \cite{yang2022towards}, and also with state-of-the-art transferable generative attack methods, including GAP  \cite{poursaeed2018generative}, CDA  \cite{naseer2019cross}, LTP \cite{salzmann2021learning}, BIA \cite{zhang2022beyond} and PDCL-Attack \cite{yang2025prompt}. These methods are all re-trained by attacking IDE \cite{zheng2016person} on DukeMTMC \cite{ristani2016performance}.} Unlike other methods, MetaAttack \cite{yang2022towards} method incorporates the color attack in addition to the additive perturbation. For a fair comparison, we only compare the attack performances of its additive perturbation. Meanwhile, based on these original methods, we train MetaAttack*, Mis-Ranking*, MUAP* and MTGA* in the ensemble training setting by attacking models in the model zoo (\textit{i.e.}, IDE \cite{zheng2016person}, PCB \cite{sun2018beyond} and ViT \cite{he2021transreid}) with dataset in data zoo (\textit{i.e.},DukeMTMC \cite{ristani2016performance}, CUHK03 \cite{li2014deepreid}, and MSMT17 \cite{Wei2018Person}). The experiment details of training data, surrogate model, target data and target model are shown in Tab. \ref{tb:tb1}. The comparison results on the mAP, aAP and mDR of six black-box attack settings are shown in Tab. \ref{tb:tb2} to Tab. \ref{tb:tb7}.

\textcolor{black}{\textbf{Comparisons with original SOTA methods.}
It can be seen that in every black-box attack scenario, our MTGA performs much better than other SOTA methods on attacking multiple black-box re-id models. For most practical and challenging cross-model\&dataset\&test scenario, our MTGA achieves a superior performance of 18.5\% aAP and 51.3\% mDR score, which outperforms the SOTA methods by 4.3\% and 11.3\% in terms of aAP and mDR. For cross-model\&dataset attack setting, our MTGA also gets the best transferability results, surpassing others by 15.4\% and 20.0\% in terms of aAP and mDR.}
\textcolor{black}{It is noteworthy that generative attack methods, including GAP, CDA, LTP, and BIA, achieve moderate performance, likely because their classification loss functions are ill-suited for the re-id retrieval task. However, PDCL-Attack attains suboptimal adversarial transferability by leveraging a vision-language model to guide semantic disruption in images. Meta-Attack demonstrates superior cross-dataset transferability due to its incorporation of diverse datasets during training, though its performance remains limited in other scenarios. Methods such as Mis-Ranking and MUAP improve transferability by introducing multi-stage discriminator networks and model-insensitive regularization terms, achieving reasonable results. Nonetheless, these two methods neither explicitly optimize transferability as a primary objective nor comprehensively consider cross-test scenarios, resulting in performance that still falls short of our approach.}

\textbf{Comparisons with ensemble trained SOTA methods.} Although the transferability of the ensemble trained SOTA methods is better than the corresponding original methods, our MTGA still performs better than the SOTA methods that use the resources of our model zoo and data zoo for ensemble training. The superiority of our MTGA than ensemble training methods can be observed in Tab. \ref{tb:tb2} to Tab. \ref{tb:tb7}. Specifically, for complicated cross-model\&dataset and cross-model\&dataset\&test black-box attack, our MTGA surpasses them by 7.6\% and 7.6\% on mDR, respectively.
\begin{table*}[t]
  \centering
  \caption{Results of \textbf{cross-model\&dataset} attack. The best performance is in \textbf{bold}.}
  \label{tb:tb6}
  % \resizebox{1\textwidth}{!}{
    \begin{tabular}{c|ccc|cc|ccc|cc}
      \hline
      \multirow{2}{*}{ Methods } & \multicolumn{3}{c|}{ Global-based } & \multicolumn{2}{c|}{ Part-based } & \multicolumn{3}{c|}{ Attention-based } & \multirow{2}{*}{ aAP$\downarrow$ } & \multirow{2}{*}{ mDR$\uparrow$ } \\
      \cline{2-9} & BOT & LSRO & MuDeep & Aligned & MGN & HACNN & Transreid & PAT & & \\
      \hline
      None & 85.4 & 77.2 & 49.9 & 79.1 & 82.1 & 75.2 & 86.6 & 78.4 & 76.7 & - \\
      \hline
      \textcolor{black}{GAP} & \textcolor{black}{46.1} & \textcolor{black}{53.9} & \textcolor{black}{19.2} & \textcolor{black}{57.7} & \textcolor{black}{60.6} & \textcolor{black}{41.8} & \textcolor{black}{66.5} & \textcolor{black}{67.1} & \textcolor{black}{51.6} & \textcolor{black}{32.7} \\
      \textcolor{black}{CDA} & \textcolor{black}{46.8} & \textcolor{black}{55.9} & \textcolor{black}{20.3} & \textcolor{black}{58.5} & \textcolor{black}{62.3} & \textcolor{black}{46.5} & \textcolor{black}{69.0} & \textcolor{black}{70.1} & \textcolor{black}{53.7} & \textcolor{black}{30.0} \\
      \textcolor{black}{BIA} & \textcolor{black}{49.9} & \textcolor{black}{60.3} & \textcolor{black}{33.9} & \textcolor{black}{61.9} & \textcolor{black}{69.8} & \textcolor{black}{59.0} & \textcolor{black}{78.5} & \textcolor{black}{66.1} & \textcolor{black}{59.9} & \textcolor{black}{21.9} \\
      \textcolor{black}{LTP} & \textcolor{black}{45.3} & \textcolor{black}{61.3} & \textcolor{black}{32.7} & \textcolor{black}{60.7} & \textcolor{black}{67.1} & \textcolor{black}{52.6} & \textcolor{black}{69.8} & \textcolor{black}{68.7} & \textcolor{black}{57.3} & \textcolor{black}{25.3} \\
      \textcolor{black}{PDCL-Attack} & \textcolor{black}{28.7} & \textcolor{black}{36.0} & \textcolor{black}{14.4} & \textcolor{black}{40.8} & \textcolor{black}{49.7} & \textcolor{black}{28.1} & \textcolor{black}{61.4} & \textcolor{black}{50.8} & \textcolor{black}{38.7} & \textcolor{black}{49.5} \\
      \hline
      MetaAttack & 26.3 & 68.6 & 37.8 & 59.4 & 73.0 & 63.9 & 80.0 & 67.7 & 59.6 & 22.3 \\
      Mis-Ranking & 46.3 & 36.7 & 11.9 & 47.5 & 46.7 & 27.0 & 65.2 & 63.4 & 43.1 & 43.8 \\
      MUAP & 42.9 & 35.7 & 9.7 & 48.0 & 40.6 & 23.8 & 58.3 & 59.7 & 39.8 & 48.1 \\
      \hline
      MetaAttack* & 38.5 & 36.5 & 18.3 & 38.0 & 44.0 & 32.6 & 62.7 & 55.0 & 40.7 & 46.9 \\
      Mis-Ranking* & 33.9 & 23.0 & 11.2 & 36.5 & 32.3 & 18.1 & 47.6 & 48.6 & 31.4 & 59.1 \\
      MUAP* & 28.7 & 19.5 & 10.3 & 36.0 & 28.5 & 20.4 & 44.0 & 45.6 & 29.1 & 62.0 \\
      \hline
      MTGA* & 31.1 & 21.8 & 8.8 & 31.3 & 27.8 & 13.8 & 42.6 & 43.6 & 27.6 & 64.0 \\
      MTGA(Ours) & \textbf{24.3} & \textbf{14.2} & \textbf{6.2} & \textbf{27.7} & \textbf{24.0} & \textbf{11.5} & \textbf{37.9} & \textbf{40.5} & \textbf{23.3} & \textbf{69.6} \\
      \hline
  \end{tabular}
  % }
\end{table*}

\begin{table*}[t]
  \centering
  \caption{Results of \textbf{cross-model\&dataset\&test} attack. The best performance is in \textbf{bold}.}
  \label{tb:tb7}
  % \resizebox{1\textwidth}{!}{
    \begin{tabular}{c|ccc|cc|ccc|cc}
      \hline
      \multirow{2}{*}{ Methods } & \multicolumn{3}{c|}{ Global-based } & \multicolumn{2}{c|}{ Part-based } & \multicolumn{3}{c|}{ Attention-based } & \multirow{2}{*}{ aAP$\downarrow$ } & \multirow{2}{*}{mDR$\uparrow$} \\
      \cline{2-9} & BOT & LSRO & MuDeep & Aligned & MGN & HACNN & Transreid & PAT & & \\
      \hline
      None & 32.7 & 33.5 & 25.8 & 35.3 & 35.8 & 29.0 & 56.2 & 56.0 & 38.0 & - \\
      \hline
      \textcolor{black}{GAP} & \textcolor{black}{20.5} & \textcolor{black}{26.3} & \textcolor{black}{18.0} & \textcolor{black}{27.5} & \textcolor{black}{31.3} & \textcolor{black}{19.7} & \textcolor{black}{42.8} & \textcolor{black}{44.7} & \textcolor{black}{28.9} & \textcolor{black}{24.1} \\
      \textcolor{black}{CDA} & \textcolor{black}{20.0} & \textcolor{black}{26.1} & \textcolor{black}{16.4} & \textcolor{black}{27.4} & \textcolor{black}{30.8} & \textcolor{black}{21.6} & \textcolor{black}{44.4} & \textcolor{black}{47.0} & \textcolor{black}{29.2} & \textcolor{black}{23.1} \\
      \textcolor{black}{BIA} & \textcolor{black}{20.9} & \textcolor{black}{27.1} & \textcolor{black}{21.8} & \textcolor{black}{29.2} & \textcolor{black}{31.9} & \textcolor{black}{25.1} & \textcolor{black}{48.1} & \textcolor{black}{50.6} & \textcolor{black}{31.8} & \textcolor{black}{16.2} \\
      \textcolor{black}{LTP} & \textcolor{black}{19.8} & \textcolor{black}{26.3} & \textcolor{black}{21.3} & \textcolor{black}{28.6} & \textcolor{black}{31.6} & \textcolor{black}{23.6} & \textcolor{black}{43.2} & \textcolor{black}{44.3} & \textcolor{black}{29.8} & \textcolor{black}{21.5} \\
      \textcolor{black}{PDCL-Attack} & \textcolor{black}{15.8} & \textcolor{black}{18.7} & \textcolor{black}{16.3} & \textcolor{black}{21.0} & \textcolor{black}{23.6} & \textcolor{black}{15.2} & \textcolor{black}{43.1} & \textcolor{black}{40.4} & \textcolor{black}{24.3} & \textcolor{black}{36.2} \\
      \hline
      MetaAttack & 16.4 & 30.0 & 22.5 & 28.2 & 34.1 & 26.1 & 53.6 & 50.9 & 32.7 & 13.9 \\
      Mis-Ranking & 19.1 & 16.7 & 12.1 & 20.5 & 24.3 & 15.8 & 41.1 & 46.6 & 24.5 & 35.5 \\
      MUAP & 18.3 & 14.1 & 12.4 & 22.6 & 20.1 & 15.5 & 36.1 & 43.4 & 22.8 & 40.0 \\
      \hline
      MetaAttack* & 19.1 & 21.3 & 17.5 & 23.1 & 24.9 & 19.3 & 45.4 & 45.0 & 27.0 & 28.9 \\
      Mis-Ranking* & 18.2 & 13.4 & 13.8 & 20.4 & 18.4 & 13.6 & 34.7 & 38.6 & 21.4 & 43.7 \\
      MUAP* & 18.3 & 15.2 & 13.6 & 24.6 & 21.4 & 16.1 & 38.5 & 40.8 & 23.6 & 38.0 \\
      \hline
      MTGA* & 16.1 & 13.0 & 11.9 & 20.6 & 18.5 & 11.6 & \textbf{31.2} & 39.0 & 20.2 & 46.8 \\
      MTGA(Ours) & \textbf{14.9} & \textbf{10.3} & \textbf{9.6} & \textbf{18.9} & \textbf{15.8} & \textbf{10.8} & 31.3 & \textbf{36.1} & \textbf{18.5} & \textbf{51.3} \\
      \hline
  \end{tabular}
  % }
\end{table*}
\subsection{Ablation Studies}

The ablation study results of CAS, PRE and NorMix modules are presented in Tab. \ref{tb:tb8}. The baseline model is trained without meta-learning scheme. It uses IDE (DukeMTMC) as the surrogate model and utilizes the DukeMTMC \cite{ristani2016performance} benchmark as training data to train the adversarial generator. Ablation experiments are tested on cross-model\&dataset black-box attack case.

\begin{table*}[t]
  \centering
  \caption{Performance analysis of each component in our MTGA.}
  \label{tb:tb8}
  % \resizebox{1\textwidth}{!}{
    \begin{tabular}{c|ccc|cc|ccc|cc}
      \hline \multirow{2}{*}{ Methods } & \multicolumn{3}{c|}{ Global-based } & \multicolumn{2}{c|}{ Part-based } & \multicolumn{3}{c|}{ Attention-based } & \multirow{2}{*}{ aAP$\downarrow$ } & \multirow{2}{*}{ mDR$\uparrow$}                                          \\
      \cline { 2 - 9 }                  & BOT                                  & LSRO                              & MuDeep                                   & Aligned                & MGN                   & HACNN & Transreid & PAT  &      &      \\
      \hline None                       & 85.4                                 & 77.2                              & 49.9                                     & 79.1                   & 82.1                  & 75.2  & 86.6      & 78.4 & 76.7 & -    \\
      \hline Baseline                       & 46.9                                 & 38.3                              & 18.5                                     & 53.8                   & 51.0                    & 26.7  & 68.4      & 63.1 & 45.8 & 40.2 \\
      +CAS                              & 30.9                                 & 19.4                              & 7.3                                      & 29.1                   & 28.9                  & 13.5  & 44.4      & 44.5 & 27.2 & 64.5 \\
      +PRE                              & 27.5                                 & 16.3                              & 7.7                                      & 29.1                   & 25.6                  & 13.8  & 41.8      & 42.8 & 25.5 & 66.7 \\
      +NorMix                           & 24.3                                 & 14.2                              & 6.2                                      & 27.7                   & 24.0                  & 11.5  & 37.9      & 40.5 & 23.3 & 69.6 \\
      \hline
    \end{tabular}
  % }
\end{table*}

\textbf{The effectiveness of CAS.} It can be observed that the incorporation of CAS module results in a significant decrease of 18.6\% in aAP and an increase of 23.7\% in mDR, which proves the effectiveness of proposed CAS module. The considerable increase in the transferability of the generated adversarial examples illustrates that the CAS module is able to simulate the black-box transfer-based attack tasks very well.
\begin{table*}[t]
  \centering
  \caption{Results on cross-model\&dataset attack w/ or w/o D.}
  \label{tb:tb11}
  % \resizebox{1.0\textwidth}{!}{
    \begin{tabular}{c|ccc|cc|ccc|c}
      \hline \multirow{2}{*}{ Methods } & \multicolumn{3}{c|}{ Global-based } & \multicolumn{2}{c|}{ Part-based } & \multicolumn{3}{c|}{ Attention-based } & \multirow{2}{*}{ aAP$\downarrow$ }                                                                                                                             \\
      \cline { 2 - 9 }                  & BOT                                  & LSRO                              & MuDeep                                   & Aligned                & MGN                    & HACNN                  & Transreid              & PAT                    &                                              \\
      \hline None                       & 85.4                                 & 77.2                              & 49.9                                     & 79.1                   & 82.1                   & 75.2                   & 86.6                   & 78.4                   & 76.7                                     \\
      \hline 
      
      w/ D                       & \textcolor{black}{24.3}               & \textcolor{black}{14.2}            & \textcolor{black}{6.2}                    & \textcolor{black}{27.7} & \textcolor{black}{24.0} & \textcolor{black}{11.5} & \textcolor{black}{37.9} & \textcolor{black}{40.5} & \textbf{23.3} \\
      w/o D                             & 25.0                                 & 15.9                              & 7.4                                      & 29.9                   & 26.2                   & 13.4                   & 40.5                    & 43.3                  &  \textbf{25.2}                                \\
      \hline
    \end{tabular}
  % }
\end{table*}

\begin{table*}[t]
  \centering
  \caption{Results on cross-model case with ensemble attacks.}
  \label{tb:tb10}
  % \resizebox{1.0\textwidth}{!}{
      \begin{tabular}{c|ccc|cc|ccc|c}
          \hline \multirow{2}{*}{ Methods } & \multicolumn{3}{c|}{ Global-based } & \multicolumn{2}{c|}{ Part-based } & \multicolumn{3}{c|}{ Attention-based } & \multirow{2}{*}{aAcc$\downarrow$  }                                                                            \\
          \cline { 2 - 9 }                  & BOT                                  & LSRO                              & MuDeep                                  & Aligned                             & MGN                   & HACNN                 & Transreid     &PAT         & \\
          \hline
          CWA                             & 43.1                                  & 57.1                               &92.2                                     & 39.3                                & 54.5                   & 57.1                  & 43.2         & 47.9    & 54.3        \\
          AdaEA                              & 47.4                                 & 54.1                               &88.8                                     & 42.4                                 &52.6                   & 54.2                  & 49.7         & 46.3   & 54.4           \\
          NTKL                              & 68.6                                & 37.2                               &76.1                                     & 55.9                                &44.9                   & 40.4                 & 52.4      & 55.2     & 53.8           \\
          Ours                       & \textbf{33.4}                & \textbf{9.8}             & \textbf{52.0}                  & \textbf{23.2}              & \textbf{15.4} & \textbf{9.9} & \textbf{33.5}  & \textbf{32.3}  &\textbf{26.2}  \\
          \hline
      \end{tabular}
  % }
\end{table*}

\begin{figure*}[t]
  % \setlength{\abovecaptionskip}{0.2cm}
  % \setlength{\belowcaptionskip}{2cm}
  % \vspace{-1cm}
  \centering
  \subfloat[Benign images.]{
  % \subfloat[]{
    \includegraphics[width=0.21\linewidth]{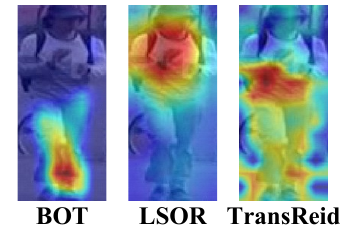}
    \label{fig:fig-4a}
  }
  \hspace{0.05\linewidth}
  \centering
  \subfloat[AE generated w/o PRE.]{
  % \subfloat[]{
    \includegraphics[width=0.21\linewidth]{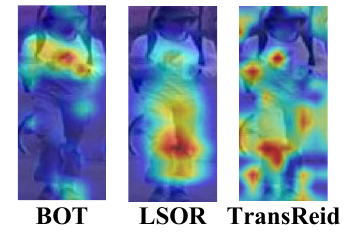}
    \label{fig:fig-4b}
  }
  \hspace{0.05\linewidth}
  \centering
  \subfloat[AE generated w/ PRE.]{
  % \subfloat[]{
    \includegraphics[width=0.21\linewidth]{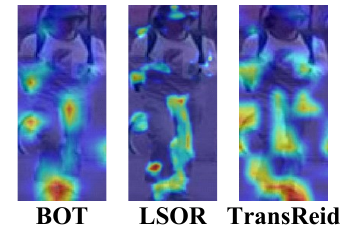}
    \label{fig:fig-4c}
  }
  % \vspace{-0.3cm}
  \caption{\textcolor{black}{Attention maps of benign images and adversarial examples (AE) on different models, visualized by Grad-CAM \cite{selvaraju2017grad}.}}
  \label{fig:fig-4}
\end{figure*}

\textbf{The effectiveness of PRE.} Tab. \ref{tb:tb8} shows the advantage of PRE module, where aAP decreases from 27.2\% to 25.5\% and mDR increases from 64.5\% to 66.7\% after the PRE module is added into the training. Also, the Grad-CAM \cite{selvaraju2017grad} visualization in Fig. \ref{fig:fig-4} shows that PRE can effectively prevent the attacker from learning to corrupt model-specific features. Concretely, Fig. \ref{fig:fig-4a} shows that models with different architectures concentrate on different part of persons. And Fig. \ref{fig:fig-4b} reflects that without the PRE module, generated adversarial examples merely mislead models to concentrate on different person part features, which results in poor transferability of attacks. Moreover, the attention maps in Fig. \ref{fig:fig-4c} demonstrate that the PRE module promotes the holistic feature corruption of person images, enhancing the transferabilities of adversarial examples.

\textbf{The effectiveness of NorMix.} The NorMix module maps the data to diverse feature subspaces, promoting the attacker to be effective not only in the feature subspace of the training models. It is seen in Tab. \ref{tb:tb8} that the NorMix module improves the mDR from 66.7\% to 69.6\%, which shows the effectiveness of our NorMix module.

\textbf{The effectiveness of discriminator.}
The discriminator is a kind of defence model that recognizes AEs generated from various domains and models, whose feedback helps attackers to generate more transferable AEs. Tab. \ref{tb:tb11} shows a degradation of attack performance without discriminator, demonstrating its effectiveness.

\textbf{The effectiveness of meta-learning.} The comparisons between MTGA (trained in meta-learning way) and MTGA* (trained in ensemble-learning way) in the Tab. \ref{tb:tb2} to Tab. \ref{tb:tb7} show that MTGA performs much better than MTGA*, which demonstrates the effectiveness of the meta-learning optimization in our method. For example, in cross-dataset and cross-dataset\&test settings, MTGA outperforms MTGA* by 5.4\% and 4.4\% mDR, respectively. The advantage of meta-learning optimization is that it learns to possess transferability capabilities by learning meta tasks, rather than get the optimal solution to the learning resources.

\begin{table}[t]
  \caption{\textcolor{black}{Comparisons on self-supervised, auxiliary feature and CLIP-based re-id models.}}
  \label{tb:tb12}
  \centering
  \begin{tabular}{c|cc|cc|cc}
      \hline
      \multirow{2}{*}{Method} & \multicolumn{2}{c|}{PASS} & \multicolumn{2}{c}{PGFA} & \multicolumn{2}{|c}{CLIP-ReID} \\ 
      \cline{2-7}
       & mAP & Rank-1              & mAP & Rank-1       & mAP & Rank-1          \\ 
      \hline
      None                    & 92.2 & 96.3               & 37.3 & 51.4          & 89.6 & 95.5         \\ 
      MTGA                    & 15.9 & 16.4               & 5.4  & 6.5         & 46.8  & 53.7          \\ 
      \hline
  \end{tabular}
\end{table}

\begin{figure}[t]
  \vspace{-0.2cm}
  \centering
  \includegraphics[width=1\linewidth]{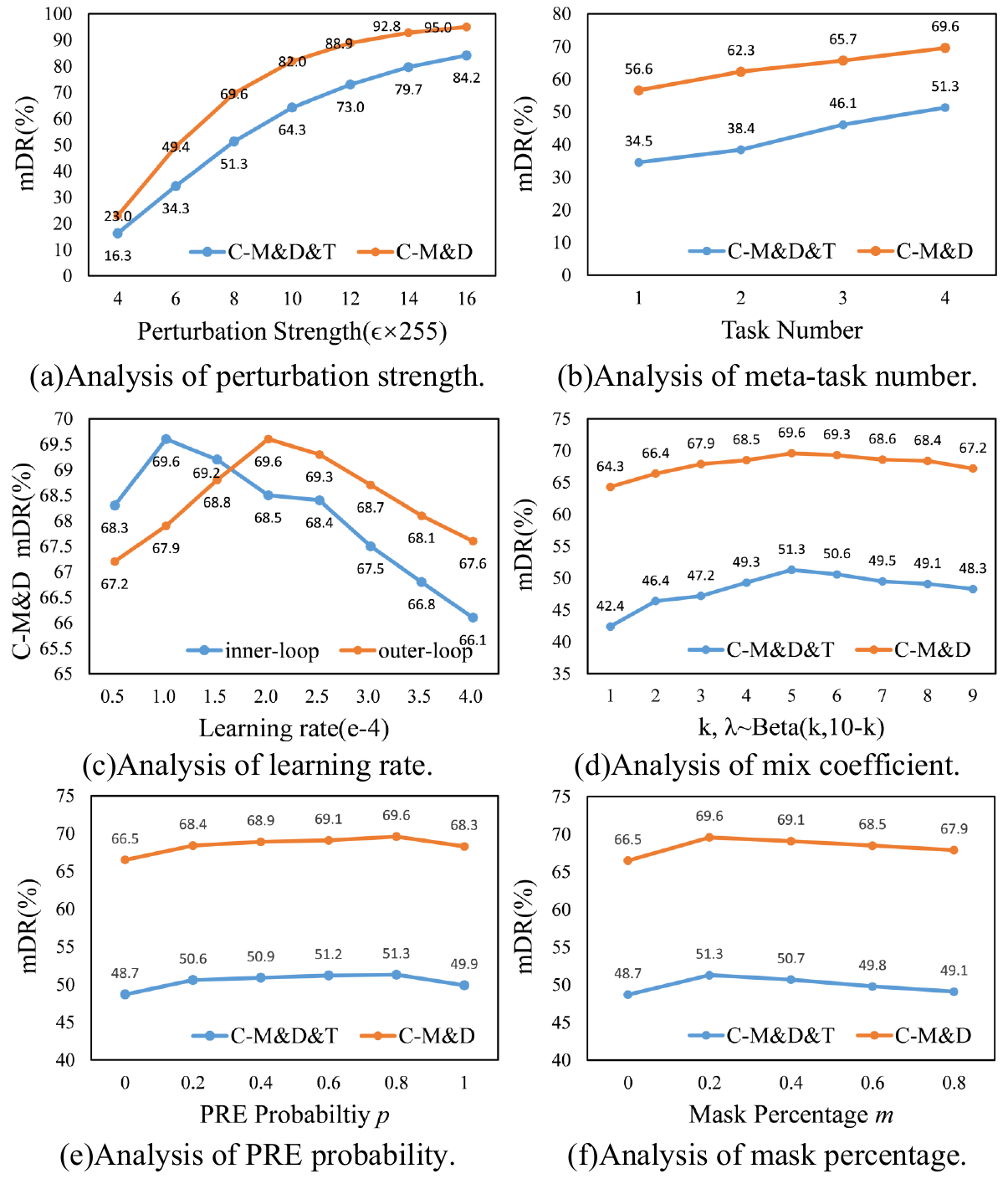}
  \caption{Analysis of mDR under different perturbation strength, task number, learning rate, mix coefficient, PRE probability and mask percentage values on the cross-model\&dataset(C-M\&D) and cross-model\&dataset\&test(C-M\&D\&T) scenarios.}
  \label{fig:parameter}
\end{figure}

To further verify the effects of meta-learning and eliminate the effects of data zoo and model zoo, we compare with SOTA classification ensemble attacks (\textit{i.e.}, CWA \cite{chen2023rethinking}, AdaEA \cite{chen2023adaptive}, NTKL \cite{weng2023exploring}). Since they only integrate multiple models without using multiple datasets, we retrained a model without the data zoo for fair comparison. As their adversarial instance perturbations cannot migrate to unseen query data, we compare the training data classification accuracy (Acc) in the cross-model setting. The results of them using the same model zoo in Tab. \ref{tb:tb10} show our method's superiority and meta-learning's effectiveness. 

\textcolor{black}{\textbf{Transferability to diverse types of re-id models.} To assess MTGA's transferability to model types beyond those in the model zoo, we conducted experiments on self-supervised PASS (Market) \cite{zhu2022pass} model, auxiliary-feature-enhanced PGFA (Occ-Duke) \cite{miao2019pose} model and CLIP-based CLIP-ReID (Market) \cite{li2023clip} model. As shown in Tab. \ref{tb:tb12}, MTGA significantly reduces the performance of all models, which demonstrates the MTGA's effectiveness against diverse model types.}

\textbf{Analysis of key parameters.} We conducted experiments on different perturbation strength, task number, learning rate, mix coefficient, PRE probability and mask percentage values. Fig. \ref{fig:parameter} shows the mDR under different settings. Larger values for perturbation strength and task number generally improve transferability, but we chose 8/255 and 5, respectively, to balance imperceptible perturbations and GPU memory. For mix coefficient and learning rate, MTGA demonstrates strong generalization and stability across different values.
\textcolor{black}{For PRE probability and mask percentage, the value 0 indicates that no PRE policy is performed and the attack performance is modest. When it is not 0, the effect is improved, and we choose the best parameters 0.8 and 0.2 of $p$ and $m$ as the experimental parameters.}
\begin{table}[t]
  \centering
  \vspace{-0.1cm}
  \caption{Results of SSIM on DukeMTMC.}
  % \resizebox{0.5\textwidth}{!}{
      \begin{tabular}{c|cccc}
          \hline Methods  & MetaAttack & MUAP   & Mis-Rank  & Ours  \\
          \hline SSIM     & 0.838     & 0.948 & 0.951       & \textbf{\textcolor{black}{0.935}} \\
          \hline
      \end{tabular}
  % }
  \label{tb:tb9}
\end{table}
\begin{figure}[t]

  \vspace{-0.2cm}
  \centering
  \subfloat{
    \includegraphics[width=0.42\linewidth]{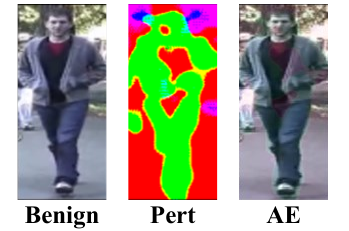}
    % \label{fig:fig-5a}
  }
  \hspace{0.05\linewidth}
  \centering
  \subfloat{
    \includegraphics[width=0.42\linewidth]{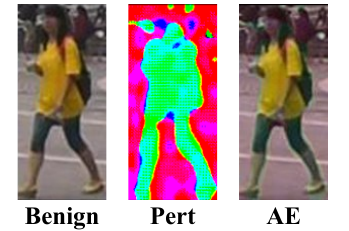}
    % \label{fig:fig-5b}
  }
  \hspace{0.05\linewidth}
  \centering
  \subfloat{
    \includegraphics[width=0.42\linewidth]{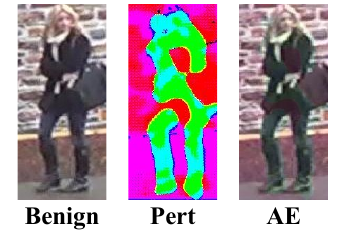}
    % \label{fig:fig-5c}
  }
  \hspace{0.05\linewidth}
  \centering
  \subfloat{
    \includegraphics[width=0.42\linewidth]{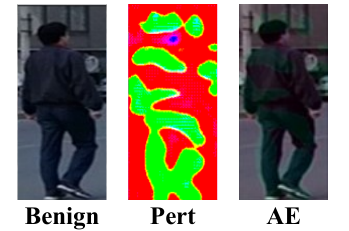}
    % \label{fig:fig-5d}
  }

  \caption{\textcolor{black}{Visualization of perturbations (Pert) and adversarial examples (AE) that generated by our MTGA across multiple datasets. The perturbations are imperceptible and human body-like.}}
  \label{fig:fig-5}
\end{figure}

\subsection{Adversarial Example Quality}
To evaluate the image quality for generated adversarial examples, we compare the SSIM \cite{wang2004image} with other attack methods for re-id. SSIM calculates structural similarity between synthetic and natural images and larger SSIM scores indicate better quality of synthetic images. 
The results of SSIM between AEs($\epsilon$=8/255) and benign images on DukeMTMC are show in Tab. \ref{tb:tb9}, which shows that our MTGA can obtain AEs with comparable quality.
\begin{figure*}[t]
  \centering
  \includegraphics[width=0.8\linewidth]{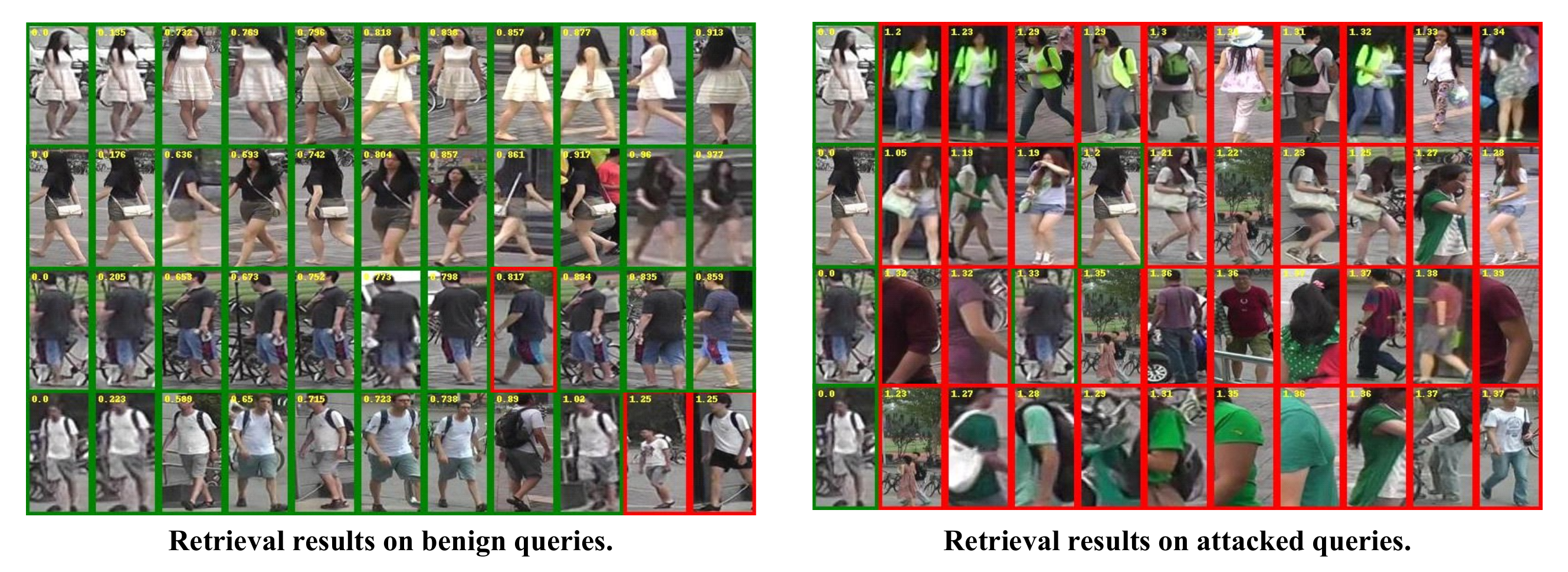}
  \caption{\textcolor{black}{The rank-10 retrieval results of BOT (Market) before and after our attack on Market.}}
  \label{fig:SupFig3-bot}
\end{figure*}

\begin{figure*}[t]
  \centering
  \includegraphics[width=0.8\linewidth]{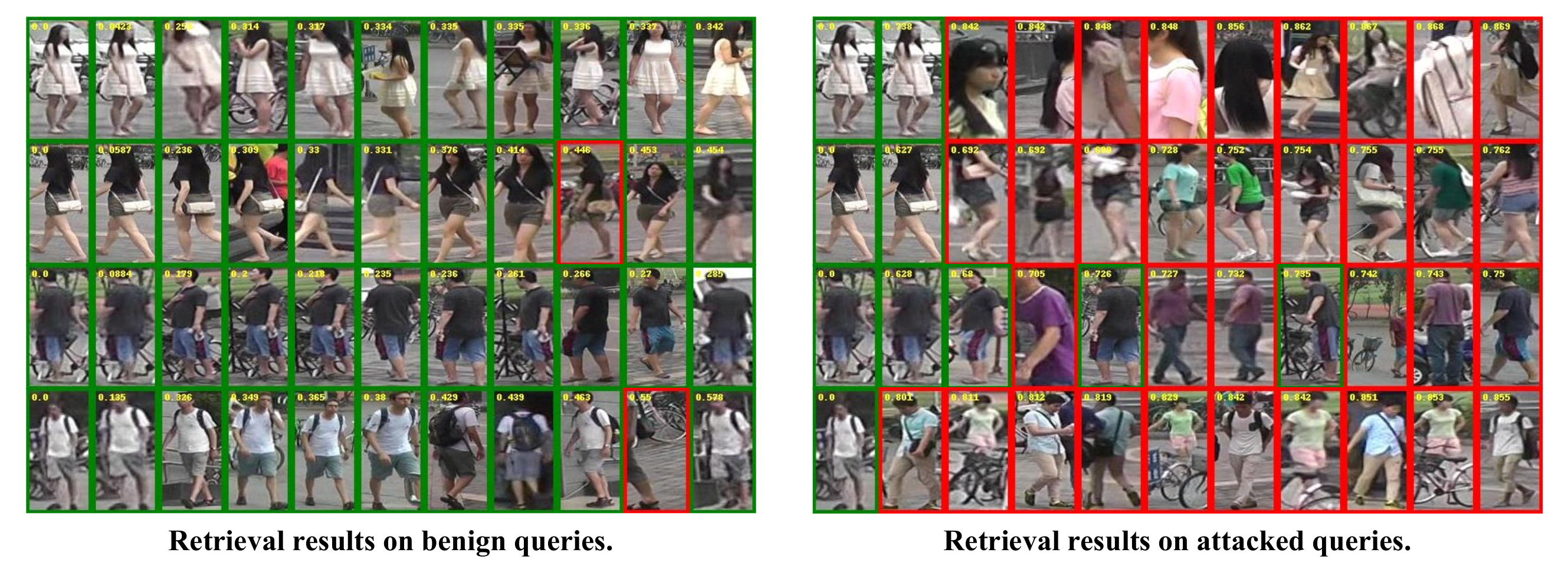}
  \caption{\textcolor{black}{The rank-10 retrieval results of Transreid (Market) before and after our attack on Market.}}
  \label{fig:SupFig3-transreid}
\end{figure*}

\subsection{Visualization}
\textcolor{black}{We visualize the perturbations and adversarial examples generated by our MTGA across multiple datasets, including  Market \cite{zheng2015scalable}, DukeMTMC \cite{ristani2016performance}, MSMT17 \cite{Wei2018Person} and VIPeR \cite{gray2007evaluating}. As Fig. \ref{fig:fig-5} shows, the perturbations on adversarial examples are imperceptible. It's hard for humans to detect the maliciously attacked adversarial examples generated by our MTGA. What's more, the generated perturbations obtain the human shape of benign images, which indicates that our MTGA is able to understand the target that needs to be attacked and attempts to perform a full range of feature destruction for different style person images, thus generating more generic adversarial attacks.}

\textcolor{black}{We also provide visualization of cross-model\&dataset attack results by showcasing the Rank-10 matches from the target re-id models (\textit{i.e.}, BOT (Market) \cite{luo2019bag} and TransReID (Market) \cite{he2021transreid}) before and after applying our proposed adversarial attack on Market dataset \cite{zheng2015scalable}, as illustrated in Fig. \ref{fig:SupFig3-bot} and Fig. \ref{fig:SupFig3-transreid}. In these figures, green boxes denote correctly matched images, red boxes indicate mismatched images, and the first column represents the query images. These visualizations demonstrate the effectiveness of our method in attacking various re-id models.}

\begin{table}[t]
  \centering
  \caption{\textcolor{black}{Attack effectiveness against defense methods}}
  \begin{tabular}{c|ccc|c|c}
    \hline Method & Adv.Res & Randomization & JPEG & aAP$\downarrow$ & mDR$\uparrow$ \\
    \hline None & 69.6 & 84.6 & 83.8 & 80.0 & - \\
    \hline MetaAttack & 67.1 & 67.8 & 57.9 & 64.3 & 19.7 \\
    Mis-Ranking & 56.1 & 43.3 & 51.2 & 50.2 & 37.3 \\
    MUAP & 53.6 & 48.5 & 57.4 & 53.2 & 33.5 \\
    \hline MTGA(Ours) & 40.3 & 26.3 & 31.8 & 32.8 & 59.0 \\
    \hline
    \end{tabular}
  \label{tab:defense_results}
\end{table}
\begin{table}[t]  

  \caption{\textcolor{black}{Comparison of Computation Cost and Performance}}
  \label{tab:tab10}
  \centering
  \resizebox{0.49\textwidth}{!}{  
  \begin{tabular}{c|ccc|c|c}
      \hline Methods                   & Parameters           & FLOPs  & Training Time           & aAP$\downarrow$            & mDR$\uparrow$  \\
      \hline
      Baseline          & $8.419\mathrm{K}$ & $180.355\mathrm{M}$ & $4.0\mathrm{h}$ & $45.8$ & $40.2$   \\
      MTGA              & $8.419\mathrm{K}$  & $180.355\mathrm{M}$ & $11.2\mathrm{h}$  & $23.3$ &$69.6$ \\
      \hline
    \end{tabular}
  }
\end{table}
\subsection{Attack Effectiveness against Defense Method} 
\textcolor{black}{In Tab. \ref{tab:defense_results}, we present additional evaluations to assess the effectiveness of our method against various defense strategies, including adversarially trained models (Adv. Res \cite{bouniot2020vulnerability}), input preprocessing techniques (JPEG compression \cite{das2017keeping}), and denoising-based methods (Randomization \cite{xie2018mitigating}). For the JPEG defense, a compression rate of 60\% was applied, and the victim model used for evaluation was BOT (Market) \cite{luo2019bag}. Our method consistently demonstrates superior attack effectiveness across these defenses. Notably, against these three categories of defenses, our approach achieves an mDR of 59.0\%, highlighting the pressing need for more robust defense mechanisms to ensure the security of re-ID systems.}

\section{Discussion}
\textcolor{black}{\textbf{Computation cost.} The comparison of computational cost and transferability performance between the baseline model and our proposed MTGA is presented in Tab. \ref{tab:tab10}. For attack inference, both methods utilize the same adversarial generator, resulting in identical model parameters and FLOPs. Regarding training time, our MTGA incorporates a meta-learning scheme, which increases the training time to more than twice that of the baseline without meta-learning optimization. However, this additional training time yields a significant improvement in transferability performance. Importantly, during the testing phase, both methods maintain identical inference efficiency, ensuring no additional computational overhead.}

\textcolor{black}{\textbf{Impact.} Our proposed adversarial attack method may potentially be exploited by malicious attackers to compromise surveillance systems, which also alters the security of re-id system and provides an effective benchmark for testing the robustness of real-world models. In the future, we plan to leverage the adversarial examples proposed in this work to further explore the development of more robust re-id models.}

\textcolor{black}{\textbf{Limitation and future work.} Our method integrates data zoo and model zoo with meta-learning, which is time-consuming and resource-intensive in training process. To address this limitation, future work will explore the introduction of visual-language models, which can provide joint visual and text representations, enabling more efficient and effective feature disruption while reducing computational costs.}

\section{Conclusion}  
In this paper, we propose a novel Meta Transferable Generative Attack method to facilitate the attacker generating highly transferable adversarial examples on black-box re-id models by learning from extensive simulated transfer-based meta attack tasks. The proposed Cross-model\&dataset Attack Simulation method constructs the cross-model and cross-dataset attack tasks by selecting different model and data for meta-train and meta-test process.
PRE strategy randomly erases the generated perturbation to suppress the model-specific feature corruption. NorMix module mimics diverse feature embeddings to boost the cross-test transferability. Comprehensive experiments show the superiority of our proposed MTGA over the state-of-the-art methods.

\bibliography{egbib}{}
\bibliographystyle{IEEEtran}
\begin{IEEEbiography}
  [{\includegraphics[width=1in,height=1.25in,clip,keepaspectratio]{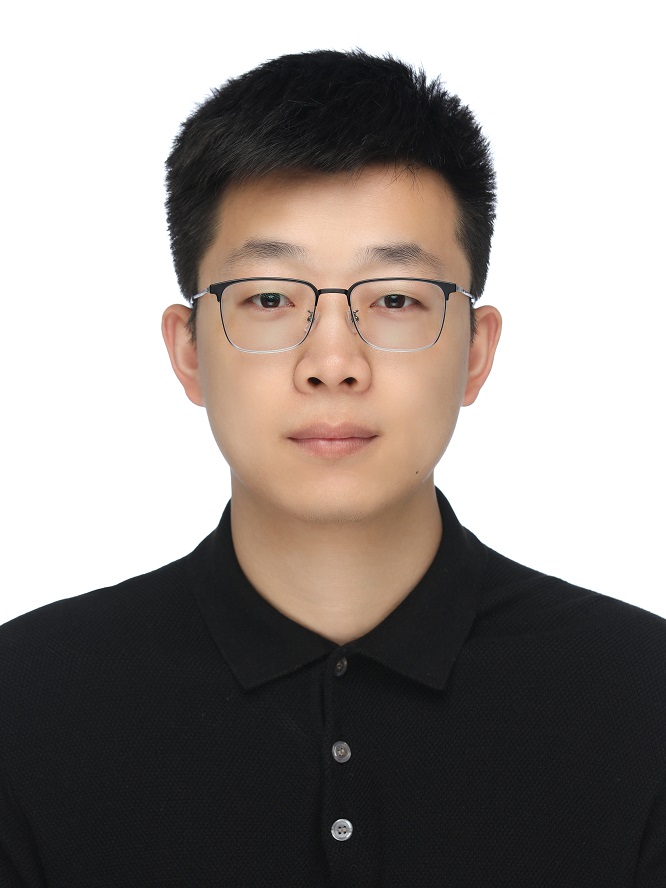}}]{Yuan Bian}
  received the master's degree from Beihang University, Beijing, China, in 2021.
  He is currently pursuing the Ph.D. degree with the College of Electrical and Information Engineering, Hunan University, Changsha, China. His research interests include computer vision, person re-identification and adversarial attack.
\end{IEEEbiography}
\begin{IEEEbiography}
  [{\includegraphics[width=1in,height=1.25in,clip,keepaspectratio]{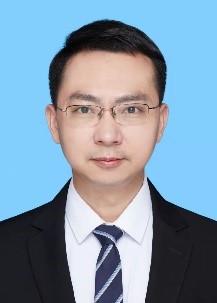}}]{Min Liu}
  received the bachelor's degree from Peking University and the PhD degree in electrical engineering from the University of California, Riverside, in 2012. He is a professor with Hunan University. He is an associate editor of IEEE Transactions on Neural Networks and Learning Systems. His research  interests include robot vision and pattern recognition.
\end{IEEEbiography}
\begin{IEEEbiography}
  [{\includegraphics[width=1in,height=1.25in,clip,keepaspectratio]{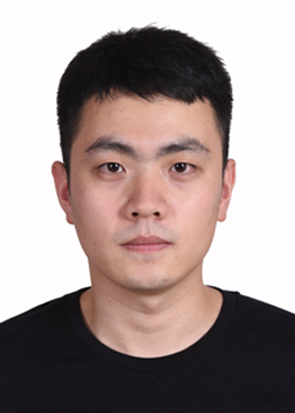}}]{Xueping Wang}
  is an assistant professor at Hunan Normal University. He received the Ph.D. degree in the College of Electrical and Information Engineering, Hunan University, China, in 2021. His research interests include computer vision, person re-identification, and adversarial attack and defense method.
\end{IEEEbiography}
\begin{IEEEbiography}
  [{\includegraphics[width=1in,height=1.25in,clip,keepaspectratio]{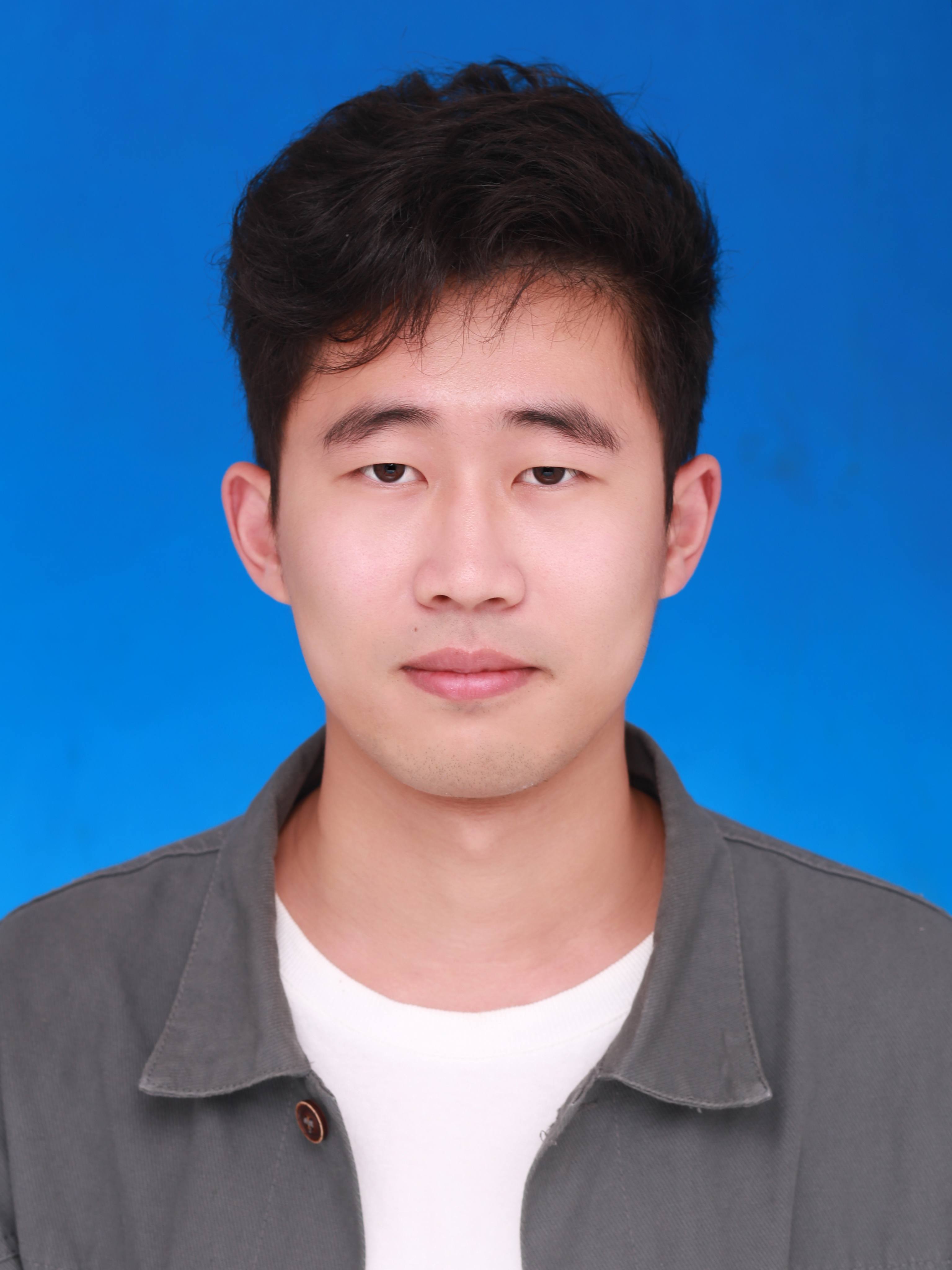}}]{Yunfeng Ma} 
  received his bachelor's degree in automation from Hunan University in 2021. He is currently pursuing the Ph.D. degree with the College of Electrical and Information Engineering, Hunan University, Changsha, China. His research interests include computer vision, anomaly detection and AutoML.
  \end{IEEEbiography}
\begin{IEEEbiography}
  [{\includegraphics[width=1in,height=1.25in,clip,keepaspectratio]{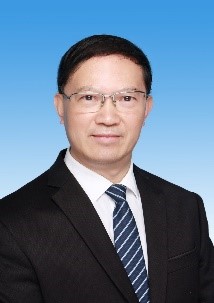}}]{Yaonan Wang}
  received the Ph.D. degree in electrical engineering from Hunan University, Changsha, China, in 1994. Since 1995, he has been a Professor with the College of Electrical and Information Engineering, Hunan University. From 1994 to 1995, he was a Post-Doctoral Research Fellow with the Normal University of Defense Technology, Changsha. From
  1998 to 2000, he was supported as a Senior Humboldt Fellow by the Federal Republic of Germany at the University of Bremen, Bremen, Germany. From 2001 to 2004, he was a Visiting Professor at the University of Bremen. He is a member of the Chinese Academy of Engineering. His research interests include robotics and image processing.
\end{IEEEbiography}
\end{document}